\documentclass{article}

\usepackage{microtype}
\usepackage{graphicx}
\usepackage[most]{tcolorbox}
\usepackage{subcaption}
\usepackage{booktabs} 
\usepackage{xcolor}

\usepackage{hyperref}

\usepackage[accepted]{icml2026}

\makeatletter
\providecommand{\ICML@appearing}{}
\makeatother

\usepackage{amsmath}
\usepackage{amssymb}
\usepackage{mathtools}
\usepackage{amsthm}
\usepackage{placeins}

\usepackage[capitalize,noabbrev]{cleveref}

\theoremstyle{plain}

\theoremstyle{definition}

\theoremstyle{remark}

\icmltitlerunning{Consensus is Not Verification}

\begin{document}

\twocolumn[
  \icmltitle{Consensus is Not Verification: Why Crowd Wisdom Strategies Fail for LLM Truthfulness}

  \icmlsetsymbol{equal}{*}

\begin{icmlauthorlist}
\icmlauthor{Yegor Denisov-Blanch}{stanford_cs,equal}
\icmlauthor{Joshua Kazdan}{stanford_stats,equal}
\icmlauthor{Jessica Chudnovsky}{stanford}
\icmlauthor{Rylan Schaeffer}{stanford_cs}
\icmlauthor{Sheng Guan}{amazon}
\icmlauthor{Soji Adeshina}{amazon}
\icmlauthor{Sanmi Koyejo}{stanford_cs}
\end{icmlauthorlist}

\icmlaffiliation{stanford_cs}{Stanford Computer Science}
\icmlaffiliation{stanford_stats}{Stanford Statistics}
\icmlaffiliation{stanford}{Stanford University}
\icmlaffiliation{amazon}{Amazon}

\icmlcorrespondingauthor{Yegor Denisov-Blanch}{ydebl@stanford.edu}
\icmlcorrespondingauthor{Joshua Kazdan}{jkazdan@stanford.edu}
\icmlcorrespondingauthor{Sanmi Koyejo}{sanmi@cs.stanford.edu}

  \icmlkeywords{Machine Learning, ICML, Verification, Polling, LLM}

  \vskip 0.3in
]

\printAffiliationsAndNotice{\icmlEqualContribution}

\begin{abstract}
Pass@$k$ and other methods of scaling inference compute can improve language model performance in domains with external verifiers, including mathematics and code, where incorrect candidates can be filtered reliably. This raises a natural question: can we similarly scale compute to elicit gains in \emph{truthfulness} for domains without convenient verification? We show that across five benchmarks and models, surprisingly, it cannot. Even at $25\times$ the inference cost of naive sampling, polling-style aggregation yields no consistent accuracy gains over single-sample baselines and often amplifies shared misconceptions. 
We find that under uncertainty, models are better at predicting what other models will say within model ensembles than at identifying what is true, revealing a separation between \emph{social prediction} and \emph{truth verification}.  Across models and benchmarks, aggregation fails to provide a robust truth signal because language model errors are strongly correlated.  The source of correlation goes beyond any individual benchmark: we show that even when conditioned on out of distribution random strings and asked to produce pseudo-random outputs, different models produce correlated outputs. Confidence-based weighting provides no benefit because self-reported confidence fails to reliably distinguish correct from incorrect answers. 
These results delineate a boundary for inference-time scaling: in verified domains, additional samples provide more candidates for a verifier to filter; in unverified domains, additional samples merely reinforce shared misconceptions.
\end{abstract}

\section{Introduction}

Inference-time scaling has emerged as a powerful alternative to parameter scaling for large language models. By allocating additional compute at test time, models can generate multiple candidate solutions and select among them, often achieving accuracy gains. Methods such as self-consistency \citep{wang2023self}, repeated sampling \citep{brown2024monkeys}, and search have proven effective in domains like mathematics and code, where correctness can be verified automatically. In these settings, aggregation amplifies capability by filtering candidates through an external verifier.

A natural question is whether aggregation can also scale \emph{truthfulness} in domains without verification. Selection must then instead rely on internal signals such as agreement, confidence, or predicted popularity. A common intuition, borrowed from the wisdom-of-crowds literature \citep{surowiecki2004wisdom}, is that aggregating many imperfect judgments should recover the truth even when individuals err.

We find that this intuition does not transfer to language models. We evaluate multiple aggregation strategies—including majority voting, confidence weighting, and the Surprisingly Popular algorithm \citep{prelec2017solution}—across several open-source models and benchmarks spanning factual knowledge, commonsense reasoning, expert-level questions, and forecasting. We find that no aggregation method consistently improves accuracy over single-sample baselines; methods that help on one benchmark often hurt on another. On forecasting questions with outcomes postdating model knowledge cutoffs, all methods perform at chance. These results do not contradict prior successes of self-consistency; they identify the regime in which those methods stop working.

This failure traces to correlated errors. Wisdom-of-crowds rests on a critical assumption: errors must be at most weakly correlated. Human crowds satisfy this condition because individuals draw on diverse experiences and information sources. Modern language models do not. Models trained on overlapping corpora and optimized for similar objectives acquire shared priors and blind spots. This parallels adversarial transferability, where different models fail in similar ways because they learn similar features \citep{goodfellow2015explaining}. When one model produces a plausible but incorrect answer, others frequently do the same. Polling does not cancel these mistakes; it amplifies shared misconceptions, producing greater confidence without greater correctness.

What is the source of these correlations? One might suspect they stem entirely from shared knowledge and misconceptions encoded in overlapping training data. To test this, we feed models randomly generated ASCII strings and ask for a forced multiple-choice answer—a setting with no ground-truth signal whatsoever. Even on this zero-knowledge baseline, different LMs exhibit correlations as high as 0.35. This rules out shared factual knowledge as the sole explanation and suggests that correlated outputs arise from aligned inductive biases and architectural similarities, therefore persist regardless of aggregation.

We diagnose these failures mechanistically. Model errors are highly correlated both within and across model families, violating the independence that crowd wisdom requires. Self-reported confidence is misaligned with correctness and instead tracks expected agreement. Surprise-based signals that power sophisticated aggregation rules do not reliably distinguish truth from consensus. At the same time, models predict collective opinion substantially better than they predict correctness, revealing a separation between \emph{social prediction} and \emph{truth verification}.

Together, these results define a boundary for inference-time scaling. Aggregation improves performance when a verifier exists, but it cannot serve as a verifier substitute. Scaling truthfulness in unverifiable domains requires external grounding or interventions that break error correlation, rather than instantiation of the bitter lesson \citep{sutton2019bitter} involving more samples from the same epistemic prior.

To summarize our conclusions:
\begin{enumerate}
    \item We show that polling-style aggregation fails to reliably improve truthfulness across verifier-absent benchmarks even at large inference-time compute.
\item We demonstrate that language model errors are strongly correlated across samples and across model families.
\item We explain why aggregation methods like confidence, predicted popularity, and surprise reflect consensus rather than correctness.
\item We introduce a simple negative control to probe the depth of inter-model correlations - random strings with forced-choice answers - in which no underlying truth exists, and show persistent above-chance agreement. This provides mechanistic evidence that correlation reflects shared priors rather than shared knowledge.
\end{enumerate}

\section{Related Work}

Inference-time scaling complements parameter scaling by allocating compute at test time. Self-consistency decoding samples multiple reasoning paths and selects answers by majority vote \citep{wang2023self}. Subsequent work showed log-linear gains with additional samples \citep{brown2024monkeys,hughes2025bestofn}, explained these gains via heavy-tailed difficulty distributions \citep{schaeffer2025monkeys}, and demonstrated that adaptive test-time compute can outperform fixed best-of-$N$ strategies \citep{snell2025scaling}. What these successes share is that answers can be verified automatically – through proof checkers or code execution.

In domains without external verifiers, selection must rely on internal signals. One proposal is to treat aggregation itself as a proxy for verification, drawing on the wisdom-of-crowds literature \citep{surowiecki2004wisdom}. Averaging independent judgments often outperforms any individual, and repeated estimates from a single person can reduce variance through an ``inner crowd'' effect \citep{vandolder2018wisdom}. Repeated samples from a language model form an inner crowd, while pooling across models forms an outer crowd, analogous to ensembles in machine learning.

Classical ensemble theory predicts gains when component errors are diverse or weakly correlated \citep{lakshminarayanan2017simple}. In binary tasks, this assumption concerns \emph{error events}, not raw agreement: errors are correlated if, conditioned on being wrong, models tend to select the \emph{same} incorrect option. Modern language models increasingly violate this condition. When different LLMs err on the same question, they often collapse onto a single wrong answer \citep{kim2025correlated}, and error correlation grows with model capability and training similarity \citep{goel2025great}. As a result, aggregation may amplify shared misconceptions rather than cancel noise.

Several aggregation methods aim to recover truth when the majority is wrong. The Surprisingly Popular (SP) algorithm compares vote shares to predicted ones, exploiting asymmetries between informed and uninformed respondents \citep{prelec2017solution}. For SP to succeed, some respondents must know the truth and anticipate the majority’s error. While SP-style aggregation has been explored for language models, it remains unclear whether LLM populations exhibit this expert-minority structure. We test both prerequisites – knowing the truth and anticipating majority error – directly.

Other aggregation schemes weight votes by self-reported confidence, assuming higher confidence implies higher accuracy. Although base language models can be calibrated \citep{kadavath2022language}, post-training often degrades this property \citep{tian2023just,leng2024taming,xiong2024can}. Sycophancy further encourages agreement over truth \citep{sharma2023sycophancy}, suggesting that confidence may track expected consensus rather than epistemic certainty.

Relatedly, work on self-correction shows that absent external feedback, models often rationalize errors rather than detect them \citep{huang2024large}. Since crowd aggregation relies on partially independent errors, shared training data and objectives pose a fundamental obstacle.

Prior work shows that inference-time compute helps scale performance when verifiers exist, and that LLM errors are often correlated. What remains unresolved is whether this correlation limits \emph{truthfulness} in verifier-absent domains.
We show that it does: correlated errors impose a structural limit that self-aggregation rules don't overcome. 

\section{Experimental Setup}
\label{sec:setup}

Crowd wisdom relies on errors that are not strongly correlated. Our experiments test this assumption by applying polling methods to repeated samples from single models (\emph{intra-model crowds}) and to ensembles pooled across models (\emph{inter-model crowds}), while measuring whether the dependence structure required for aggregation is satisfied.

\paragraph{Tasks and response formats.}
We elicit three response types: (i) a binary answer (e.g., \texttt{YES}/\texttt{NO} or \texttt{TRUE}/\texttt{FALSE}), (ii) an answer with self-reported confidence on a 0--100 scale, and (iii) a prediction of the vote share of one canonical option (e.g., the fraction of responses answering \texttt{YES} or \texttt{TRUE}).
Aggregation rules such as majority voting and the Surprisingly Popular (SP) algorithm require a finite option space with well-defined vote shares and prediction distributions. Binary questions are therefore not a requirement but the simplest setting, avoiding discretization artifacts and enabling comparison across aggregation methods. This choice limits generalization to open-ended generation, which we discuss in Section~\ref{sec:discussion}. Binary questions represent the most favorable setting for polling-style aggregation; failure to obtain reliable gains even in this simplified regime provides strong evidence against aggregation as a general mechanism for scaling truthfulness in verifier-absent domains.

We evaluate four benchmarks spanning verifier-absent regimes:
\begin{itemize}
    \item \textbf{Com2Sense} \citep{singh2021com2sense}: binary commonsense reasoning.
    \item \textbf{Humanity’s Last Exam (HLE)} \citep{phan2025humanity}: expert-level questions, restricted to those with a naturally binary answer structure.
    \item \textbf{BoolQ} \citep{clark2019boolq}: binary factual question answering.
    \item \textbf{Predict-the-Future}: a collection of forecasting questions introduced in this work, where all ground-truth outcomes were \emph{definitively resolved and verifiable at the time of dataset construction and writing}, even though the events postdate model knowledge cutoffs.

\end{itemize}


The HLE subset yields wider confidence intervals due to its small size, but accuracy remains far below the 50\% guessing baseline (5.7\%), indicating systematic attraction to incorrect answers rather than sampling noise.

\paragraph{Sampling protocol.}
For each question and response type, we collect 25 independent samples per model at temperatures $T\in\{0.7,1.0\}$.
We use moderate-to-high temperatures because crowd aggregation requires diversity; at low temperatures, repeated sampling collapses to near-deterministic outputs.
The relevant question is not whether a single greedy sample outperforms high-temperature aggregation, but whether aggregation provides gains \emph{given} the diversity required for crowd methods to apply.
Across all benchmarks, this corresponds to 50 samples per (question, model) and a total of \emph{375,000} model across experiments; we report benchmark-wise counts in Appendix~\ref{app:benchmark}.

\begin{figure*}[t]
\centering
\includegraphics[width=\linewidth]{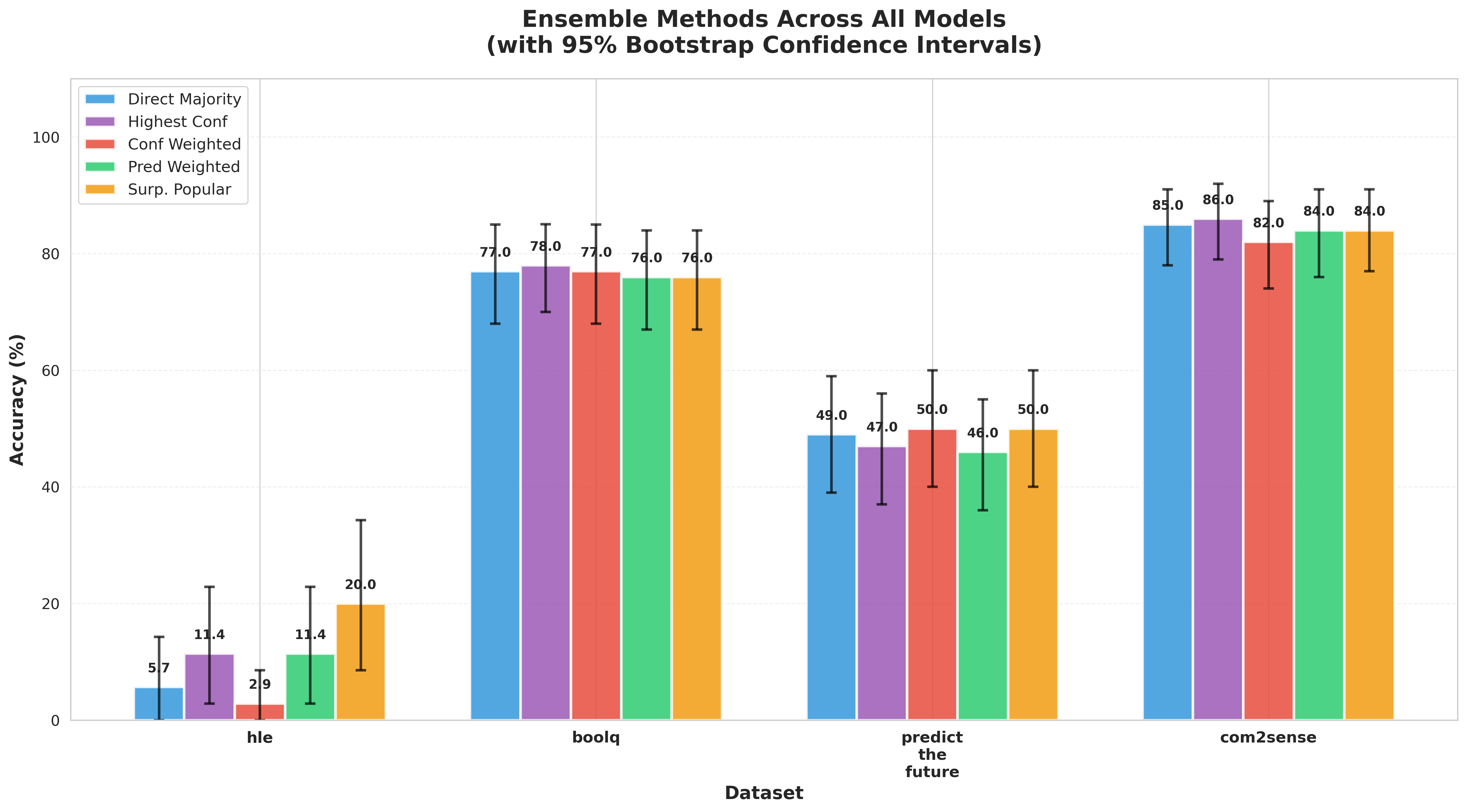}
\caption{\textbf{No ensemble aggregation method consistently outperforms majority voting across benchmarks.} We compare five aggregation rules across four benchmarks using five-model ensembles (125 votes per question). Some methods improve performance on individual tasks, but none dominates overall. Error bars show 95\% bootstrap confidence intervals.}
\label{fig:ensemble_accuracy}
\end{figure*}

\paragraph{Models and crowds.}
We evaluate five instruction-tuned, open-source models spanning 4B to 235B parameters across three families: Gemma-3-4B \citep{gemmateam2025gemma3}, GPT-oss-20B and GPT-oss-120B \citep{openai2025gptoss}, Qwen-32B \citep{qwen2024qwen}, and Qwen3-235B \citep{qwen2025qwen3}.
Intra-model crowds consist of repeated samples from a single model.
Inter-model crowds pool responses across all five models at $T=1.0$, yielding 125 votes per question.
Evaluating both regimes allows us to test whether increased architectural and training diversity reduces error correlation, as crowd wisdom would predict.

\paragraph{Aggregation methods.}
We evaluate five aggregation rules that exhaust common internal selection signals:
(1) Majority Vote \citep{decondorcet1785essay};
(2) Highest Confidence \citep{kadavath2022language};
(3) Confidence-Weighted Vote \citep{wang2023self};
(4) Prediction-Weighted Vote, weighting each response by its predicted popularity;
and (5) Surprisingly Popular (SP) \citep{prelec2017solution}, which selects the answer whose observed support exceeds its predicted support.
If self-aggregation can scale truthfulness without an external verifier, it should do so under at least one of these rules.
We additionally report an inverse-SP diagnostic solely to test alignment between the SP signal and correctness; it is not proposed as a method.

\paragraph{Evaluation and correlation.}
Performance is measured by accuracy with 95\% bootstrap confidence intervals obtained by resampling questions.
Because independence is the key assumption under test, correlation is a first-class quantity.
We measure it via majority stability under temperature variation, inter-rater reliability (including vote entropy and Fleiss’ $\kappa$), and the concentration of wrong answers when models err.
These measures distinguish genuine epistemic diversity from surface-level variation.

\paragraph{Controls and forecasting benchmark.}
To distinguish structural correlation from shared knowledge, we include control analyses in which no correct answer exists, allowing us to test whether model agreement persists even when no signal is present. Finally, we include Predict-the-Future, a forecasting evaluation set whose outcomes postdate model knowledge cutoffs. All outcomes are manually verified. If aggregation can extract truth from uncertain respondents, it should help in this setting; accuracy near chance therefore provides a stringent negative test.

\section{Results}
\label{sec:results}
Across verifier-absent benchmarks, we find that self-aggregation does not provide a reliable path from inference-time compute to truthfulness. Despite substantial increases in consensus, accuracy remains flat or degrades.
This failure is systematic rather than accidental.
We show that when language models err, their errors are often correlated, violating the independence assumptions required for crowd wisdom. Consequently, when model errors are correlated, aggregation rules based on internal signals like agreement, confidence, or predicted popularity cannot reliably scale truthfulness.


\begin{figure*}[t]
  \centering
  \begin{minipage}{0.32\textwidth}
    \centering
    \includegraphics[width=\linewidth]{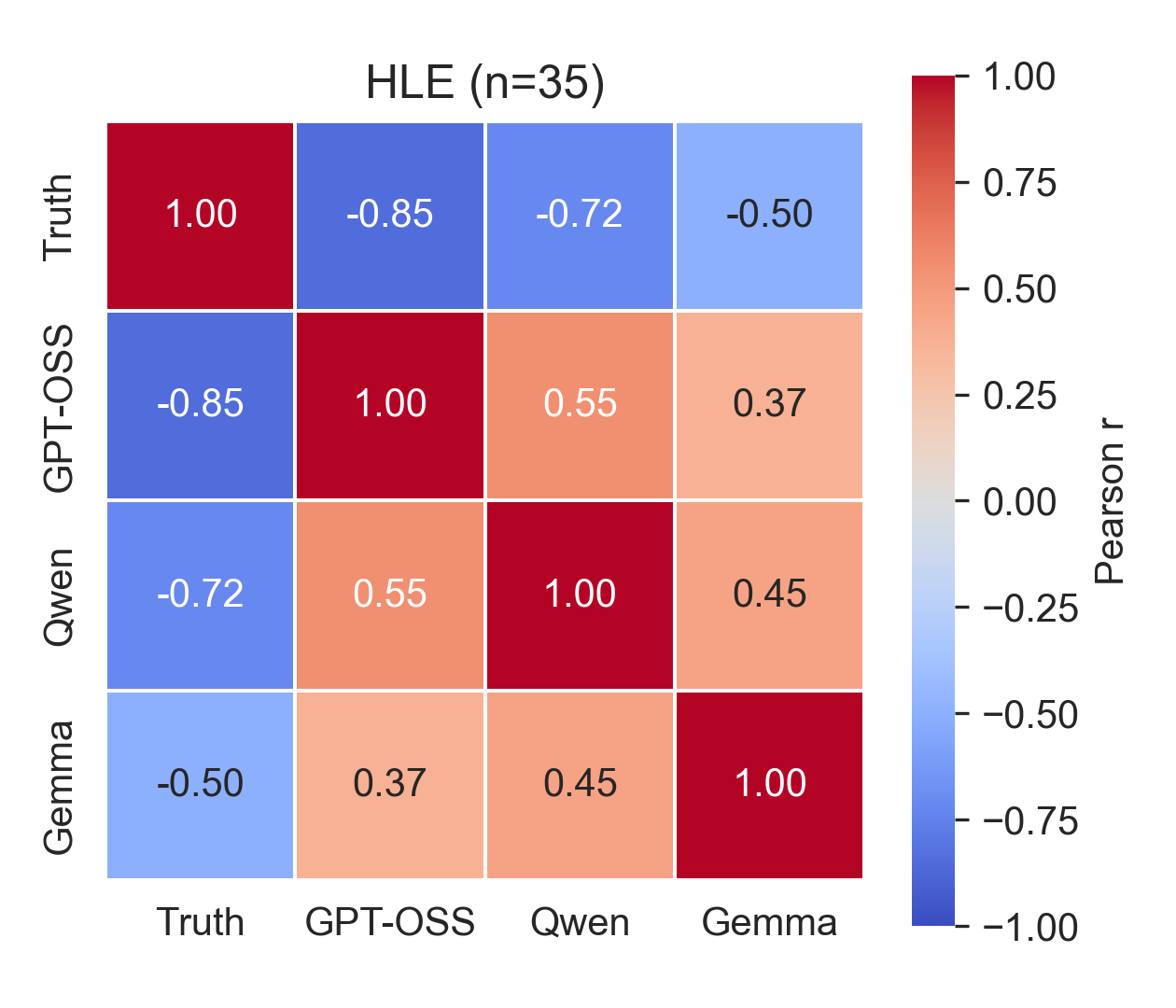}
    \subcaption{\textbf{HLE.}}
    \label{fig:corr_with_truth_hle}
  \end{minipage}
  \hfill
  \begin{minipage}{0.32\textwidth}
    \centering
    \includegraphics[width=\linewidth]{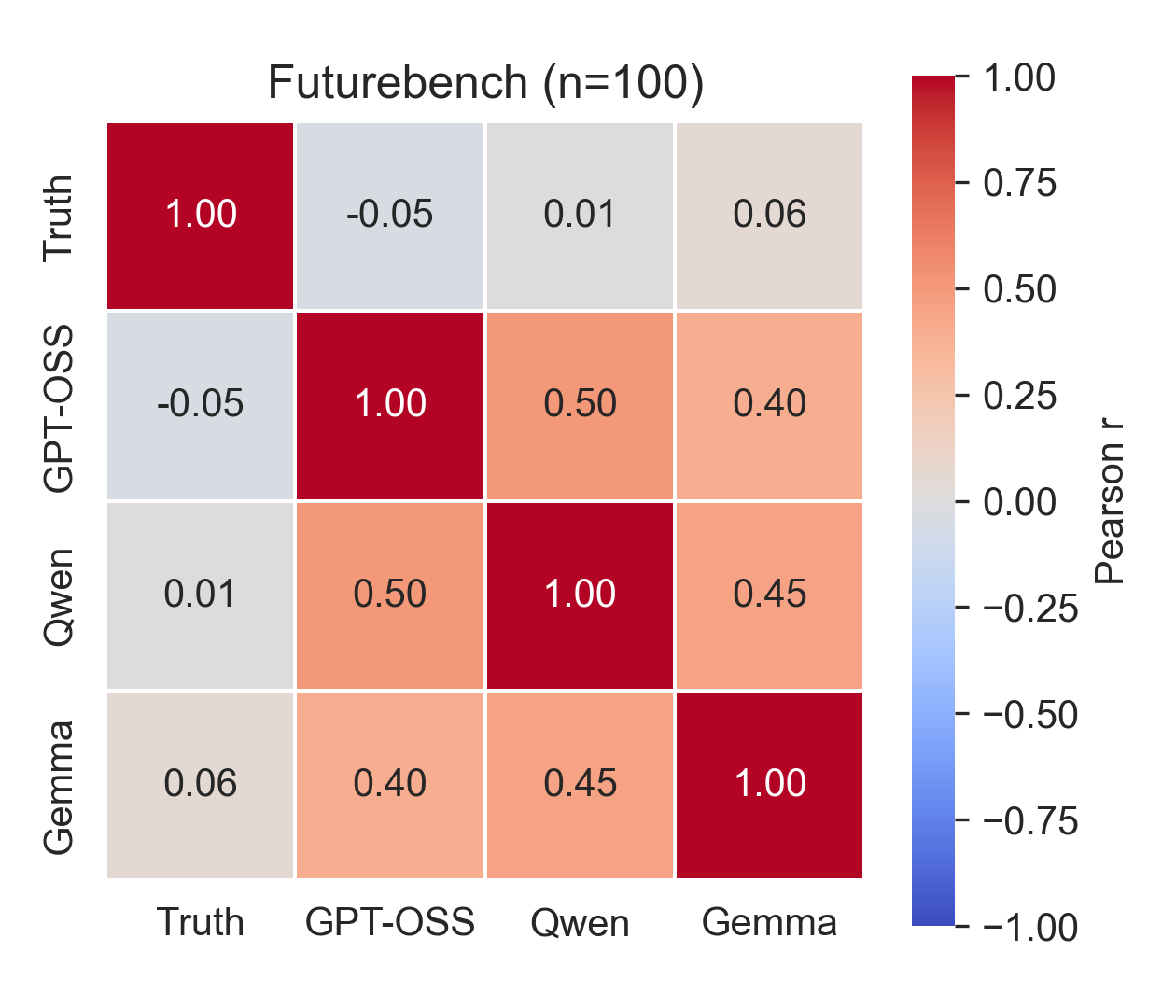}
    \subcaption{\textbf{Predict-the-Future}}
    \label{fig:corr_with_truth_future}
  \end{minipage}
  \hfill
  \begin{minipage}{0.32\textwidth}
    \centering
    \includegraphics[width=\linewidth]{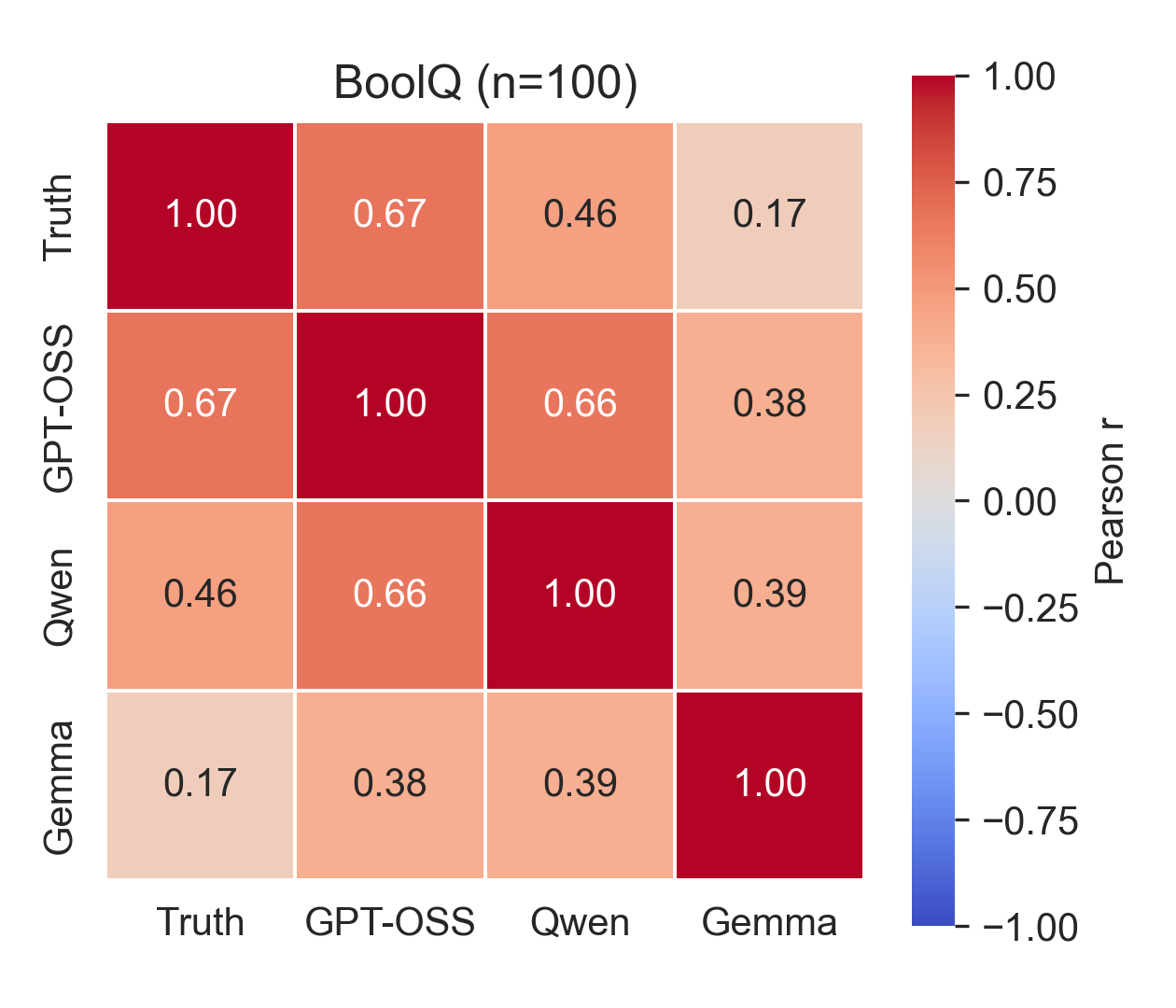}
    \subcaption{\textbf{BoolQ.}}
    \label{fig:corr_with_truth_boolq}
  \end{minipage}

  \caption{\textbf{Models agree with each other more reliably than they agree with truth.}
  Each panel reports Pearson correlations between binary answers from different model families and the ground-truth label (“Truth”), computed over questions in the benchmark.
  Across datasets, inter-model correlations are consistently positive, but correlation with Truth varies sharply by task, showing that agreement is not a stable proxy for correctness.}
  \label{fig:cross_family_answer_correlation_with_truth}
\end{figure*}

\subsection{Aggregation Fails to Improve Truthfulness}

We evaluate five aggregation rules (majority vote, highest confidence, confidence-weighted vote, prediction-weighted vote, and Surprisingly Popular) across four verifier-absent benchmarks: Humanity’s Last Exam (HLE), BoolQ, Com2Sense, and Predict-the-Future. Across models, temperatures, and datasets, no method consistently outperforms single-sample baselines (Figure~\ref{fig:ensemble_accuracy}). Increasing the number of samples increases consensus, but not correctness.

The forecasting benchmark provides the clearest negative test: because outcomes postdate model knowledge cutoffs, aggregation should succeed here if it could extract latent expertise. Instead, all methods remain indistinguishable from chance.

These failures are structural. Crowd wisdom requires errors to be uncorrelated enough that mistakes cancel under aggregation \citep{surowiecki2004wisdom,decondorcet1785essay}. Language models violate this assumption - shared training data, objectives, and post-training incentives produce shared priors and blind spots. Hence, aggregation increases consensus without increasing correctness.


Since majority voting fails, we next examine whether other model-internal signals - like confidence or predicted popularity - can recover correctness. We test confidence-based weighting and the Surprisingly Popular (SP) algorithm. Both fail because, in practice, they end up measuring expected consensus rather than truth.



Confidence does not track correctness. Self-reported confidence correlates weakly with accuracy but similarly with agreement. On hard benchmarks, confident answers are often wrong - due to sycophantic training that rewards typical-sounding outputs \citep{sharma2023sycophancy,leng2024taming}.


The SP algorithm assumes an expert minority that both knows the truth and anticipates the majority’s error \citep{prelec2017solution}.
Language model populations rarely exhibit this structure.
As a diagnostic, we evaluate inverse-SP.
On HLE, inverse-SP attains 80\% accuracy, implying that the standard SP signal is systematically anti-correlated with correctness.
On other datasets, inverse-SP performs at chance.
The sign of the surprise gap is therefore not stable across tasks, meaning it cannot serve as a reliable verification signal.


\subsection{Correlated Errors Explain the Failure}

Polling-style aggregation relies on sufficiently diverse and independent errors, under which aggregation increases the probability of correctness \citep{surowiecki2004wisdom,decondorcet1785essay}.
We test this assumption directly by measuring cross-family answer correlations and their alignment with ground truth (Figure~\ref{fig:cross_family_answer_correlation_with_truth}).

\paragraph{Correlated mistakes in verifiable domains.}  
We analyze error correlation in verifiable mathematics benchmarks - MATH \citep{hendrycks2021measuringmathematicalproblemsolving} and AIME \citep{aime} - where aggregation methods are known to help due to the presence of external verifiers.
For each dataset, we evaluate 128 problems with 200 samples per problem and focus on \emph{plurality-wrong} cases, where the most frequent answer is incorrect.

On MATH, plurality-wrong rates are low (3.9--25.8\%), but errors are highly concentrated: the most common incorrect answer accounts for 65--87\% of wrong responses. On AIME, plurality-wrong rates vary widely (3.9--75.8\%) and error concentration is substantially lower (20--63\%).

These results clarify why aggregation succeeds in mathematics. Aggregation succeeds in math because verifiers can filter out wrong answers - not because agreement signals truth. But when models converge on the same wrong answer, there is nothing correct left for the verifier to find.

\paragraph{Surprise signals do not reliably track truth.}
The Surprisingly Popular (SP) algorithm succeeds only when an expert minority both knows the truth and anticipates the majority’s error.
Using inverse-SP purely as a diagnostic, we find that the direction of the surprise gap is not stable across benchmarks.
On HLE, inverse-SP attains high accuracy, implying that the standard SP signal is systematically anti-correlated with correctness.
On other datasets, inverse-SP performs at or near chance.

The surprise gap points toward truth on some benchmarks and away from it on others. A signal that flips direction across tasks cannot serve as a verifier.

\begin{figure*}[t]
    \centering
    \begin{minipage}{0.48\textwidth}
        \centering
        \includegraphics[width=\linewidth]{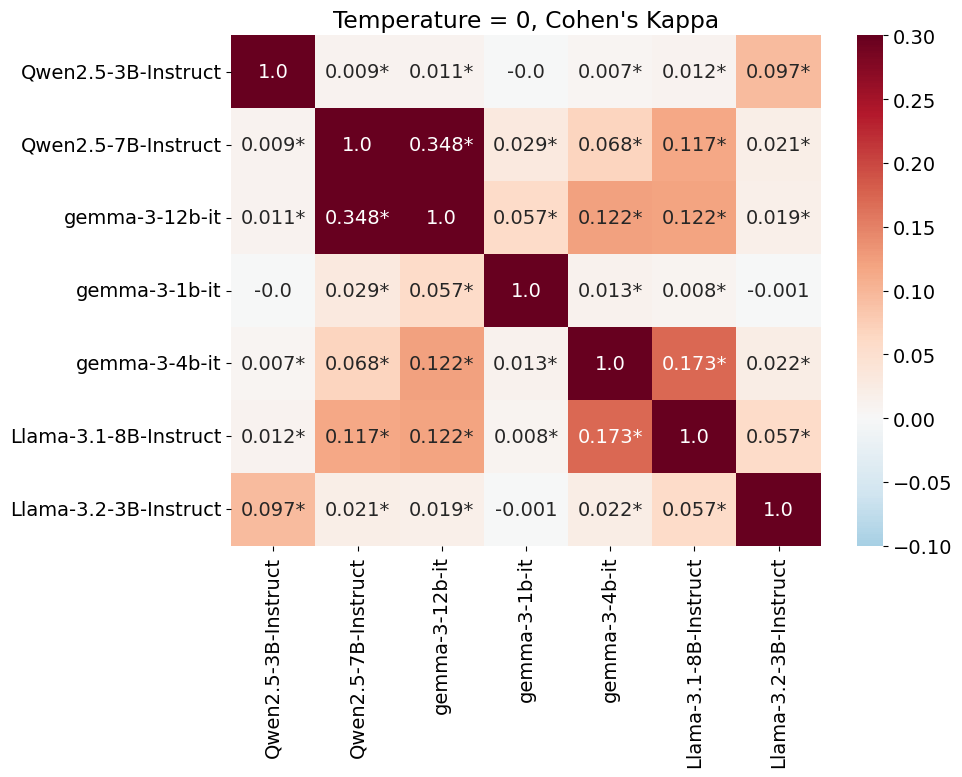}
        \subcaption{\textbf{Temp.\ 0.}}
        \label{fig:corr_no_truth_t0}
    \end{minipage}
    \hfill
    \begin{minipage}{0.48\textwidth}
        \centering
        \includegraphics[width=\linewidth]{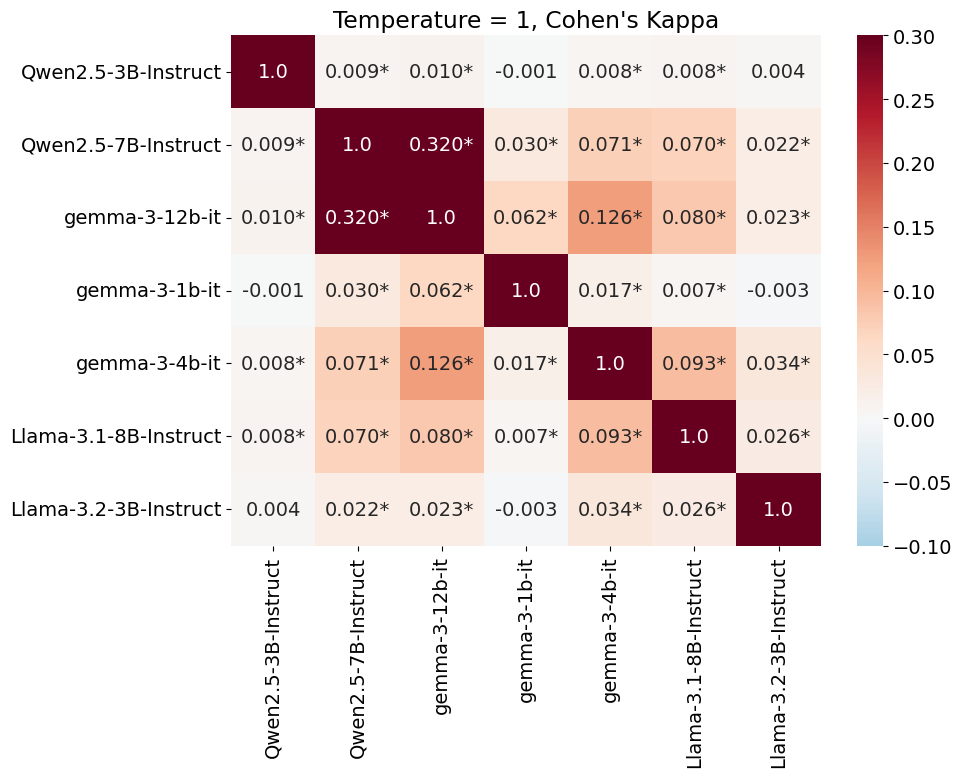}
        \subcaption{\textbf{Temp.\ 1.}}
        \label{fig:corr_no_truth_t1}
    \end{minipage}
    \caption{\textbf{Models exhibit correlated behavior even when no ground truth exists.}
    Cohen’s $\kappa$ \citep{cohen1960coefficient} between pairs of models on random strings with forced-choice answers shows stable above-chance agreement, with a similar correlation structure across temperatures, indicating shared inductive biases rather than shared knowledge.}
    \label{fig:correlation_without_truth}
\end{figure*}

\paragraph{Temperature sampling does not induce independence.}

Varying temperature has little effect on inducing diversity: between $T{=}0.7$ and $T{=}1.0$, the \emph{plurality} answer flips in only \textbf{2.9\%} of (question, model) pairs (Table~\ref{tab:temp_stability}), indicating that additional samples explore surface variation rather than distinct hypotheses.

Model ensembling likewise fails to restore independence.
Despite large differences in overall error rates across models on both MATH and AIME, incorrect answers remain highly concentrated, and on 53\% of MATH questions where multiple models err, they converge to the \emph{same} incorrect answer.

Correlation is structural, not an artifact of temperature or model choice.

\paragraph{Structural corollary.}
When errors are correlated, no aggregation rule based solely on internal signals can reliably distinguish a unanimous correct answer from a unanimous wrong one. Additional samples increase confidence without increasing correctness.

\begin{figure*}[t]
    \centering
    \begin{minipage}{0.22\textwidth}
        \centering
        \includegraphics[width=\linewidth]{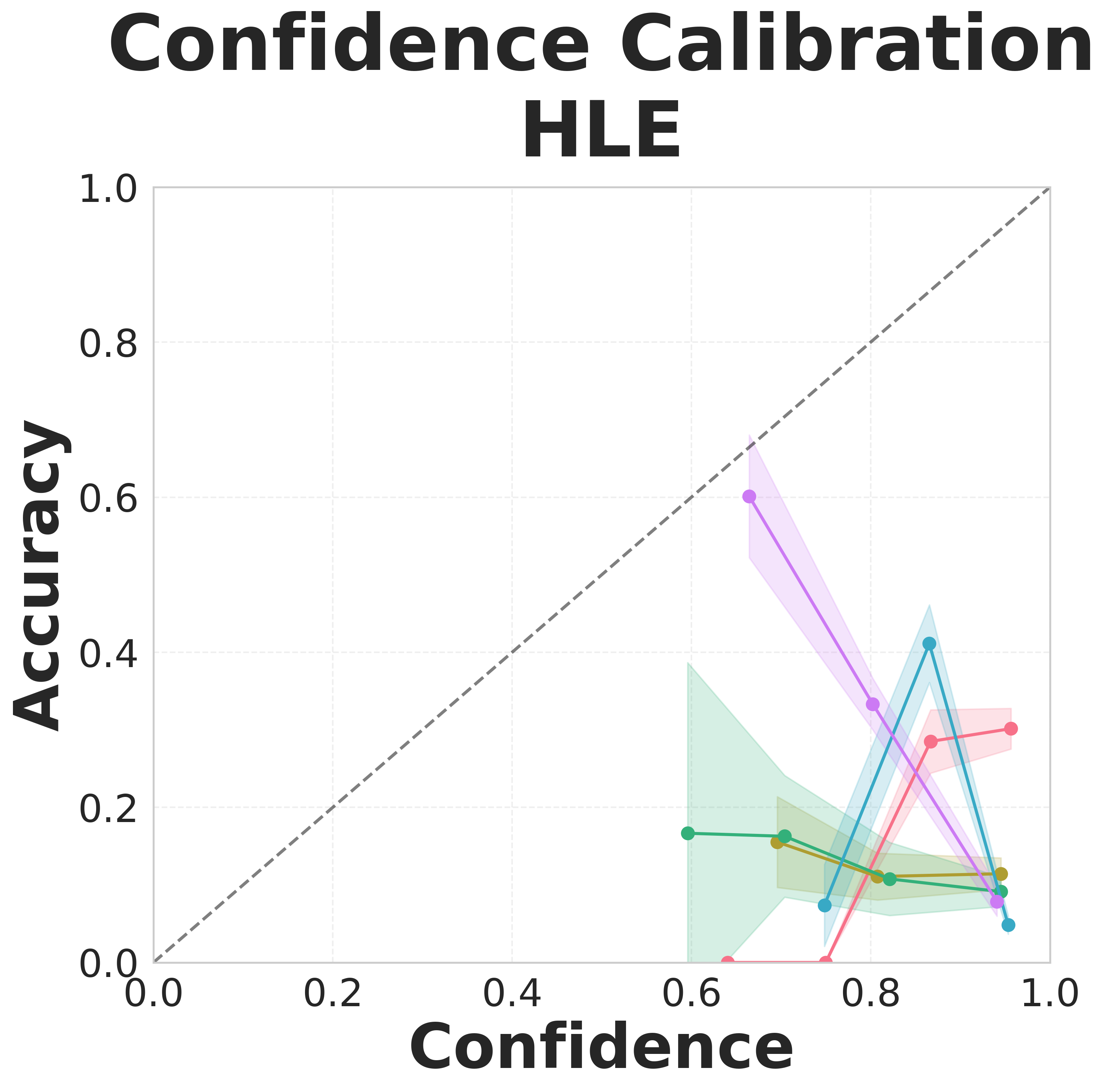}
        \subcaption{\textbf{HLE.}}
        \label{fig:calib_hle}
    \end{minipage}
    \hfill
    \begin{minipage}{0.22\textwidth}
        \centering
        \includegraphics[width=\linewidth]{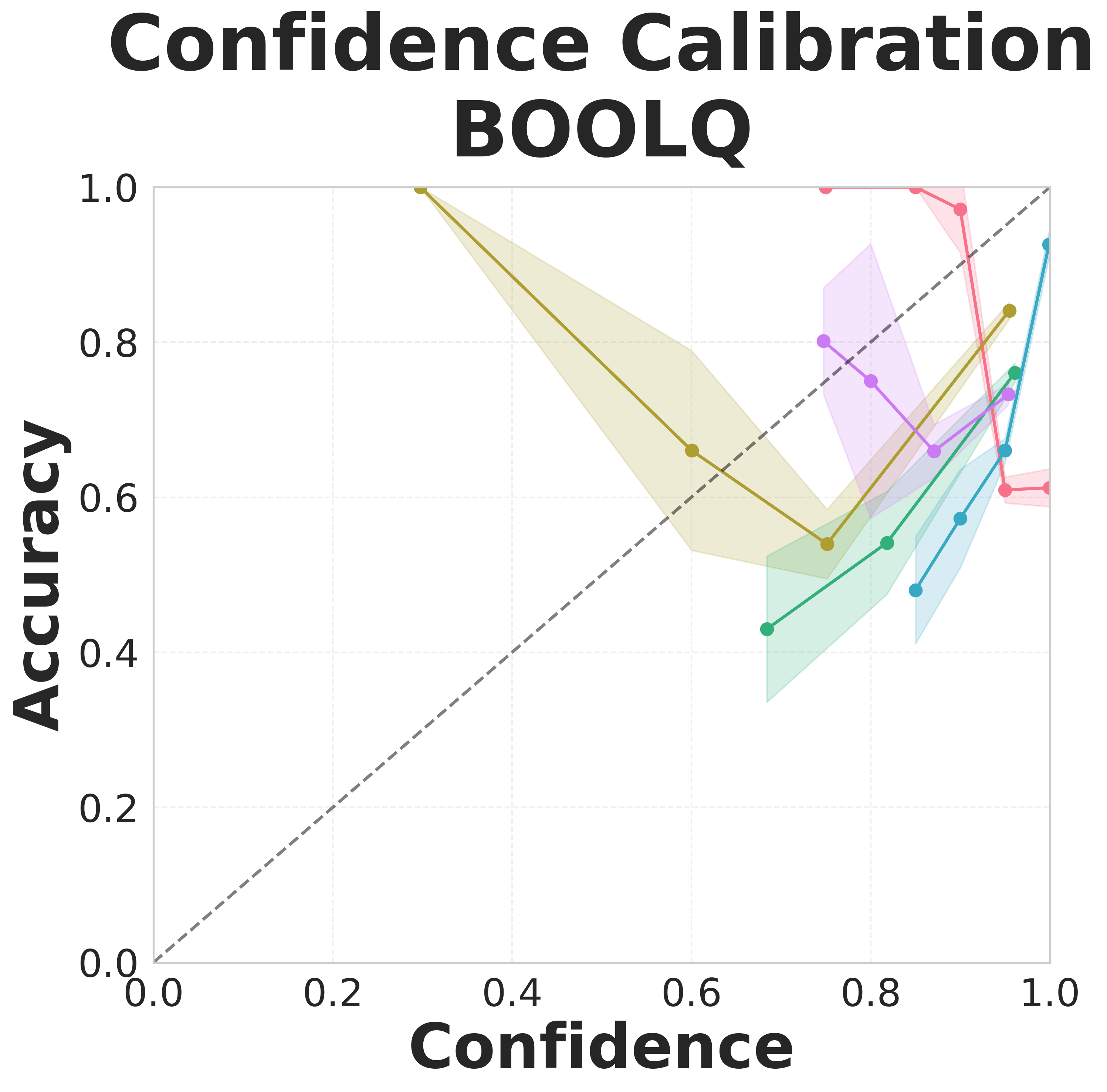}
        \subcaption{\textbf{BoolQ.}}
        \label{fig:calib_boolq}
    \end{minipage}
    \hfill
    \begin{minipage}{0.22\textwidth}
        \centering
        \includegraphics[width=\linewidth]{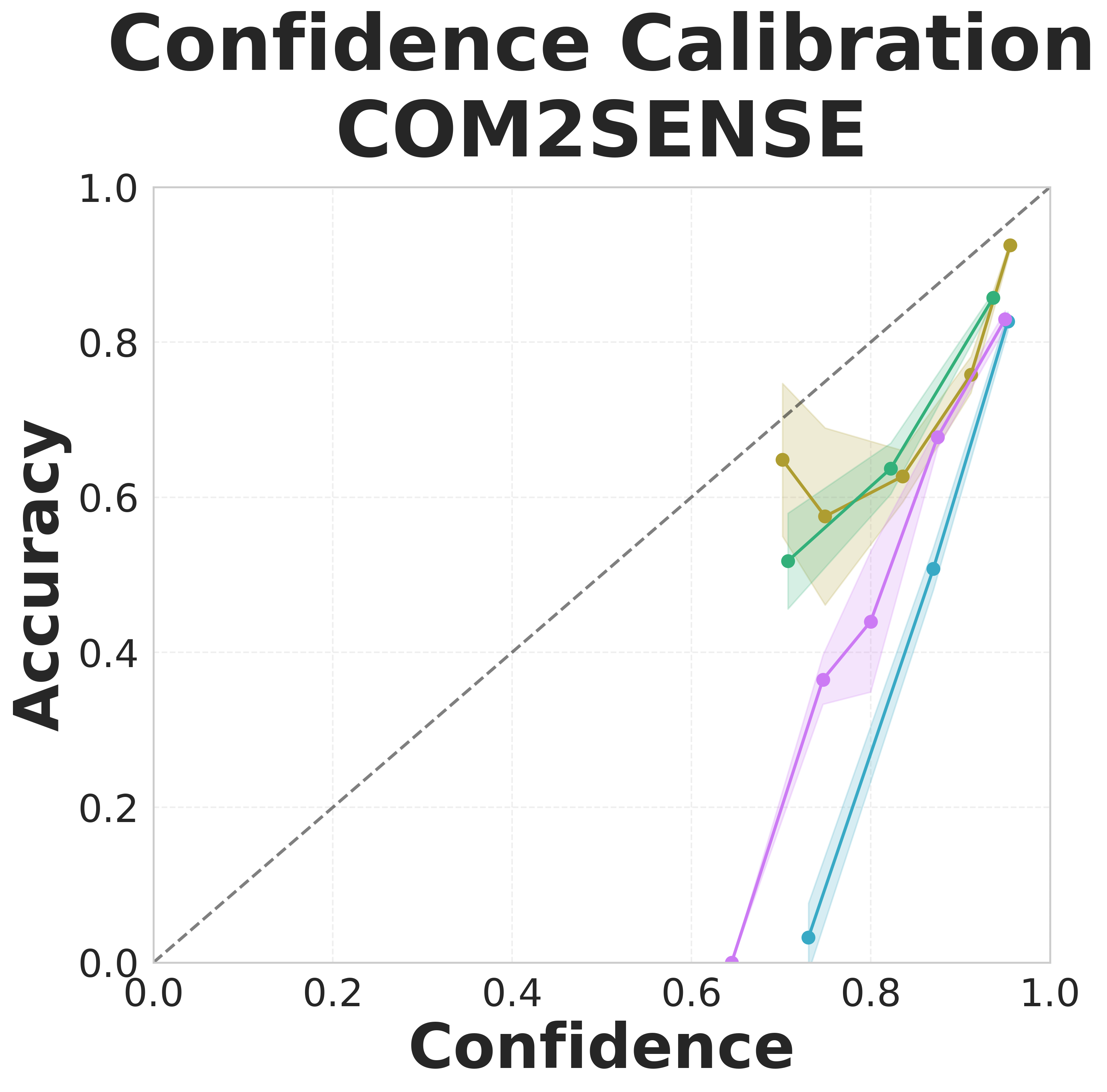}
        \subcaption{\textbf{Com2Sense.}}
        \label{fig:calib_com2sense}
    \end{minipage}
    \hfill
    \begin{minipage}{0.22\textwidth}
        \centering
        \includegraphics[width=\linewidth]{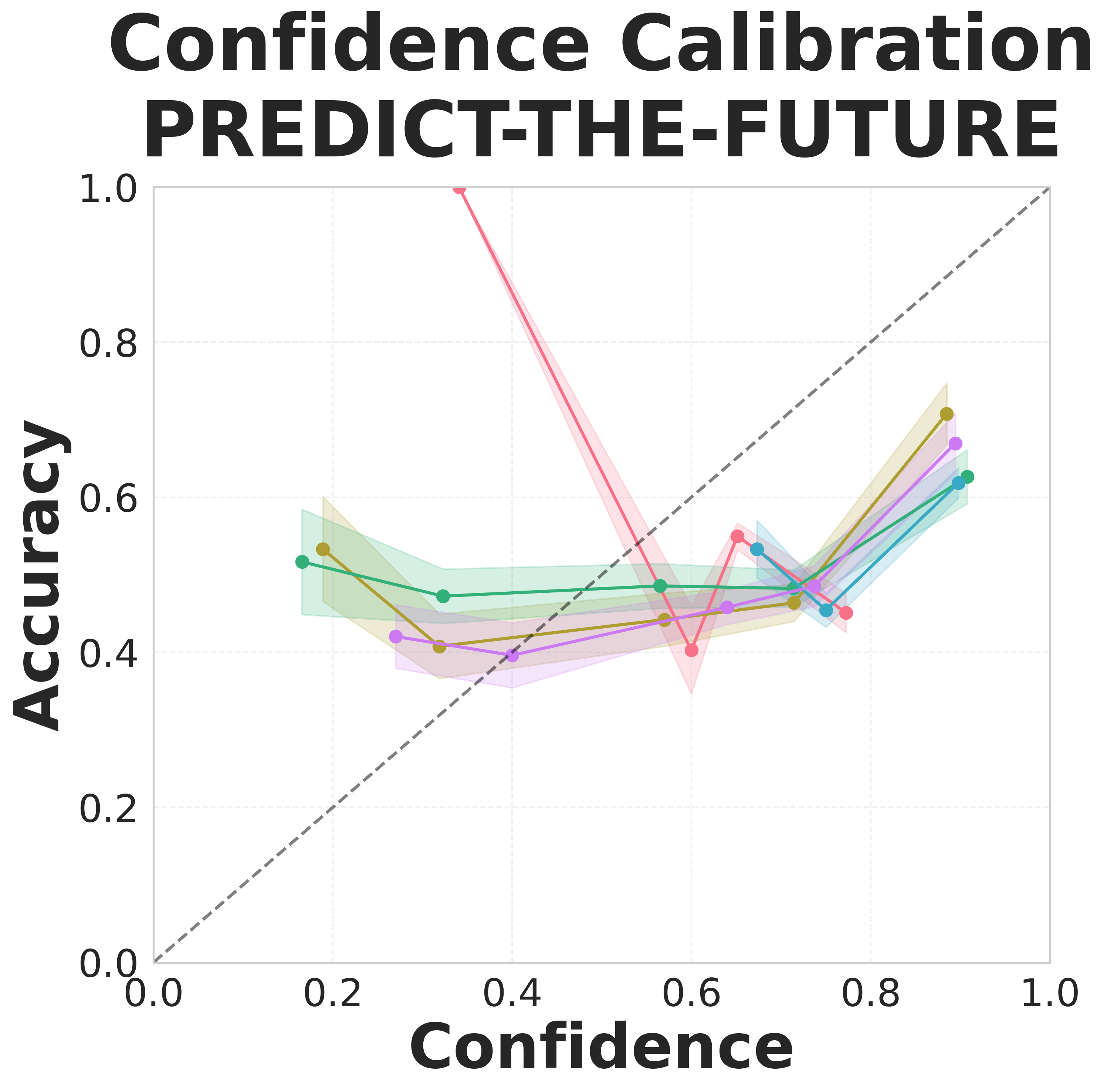}
        \subcaption{\textbf{Predict-the-Future.}}
        \label{fig:calib_future}
    
    \end{minipage}
    \hfill
    \begin{minipage}{0.10\textwidth}
      \vspace{-55pt}  
      \includegraphics[width=\linewidth]{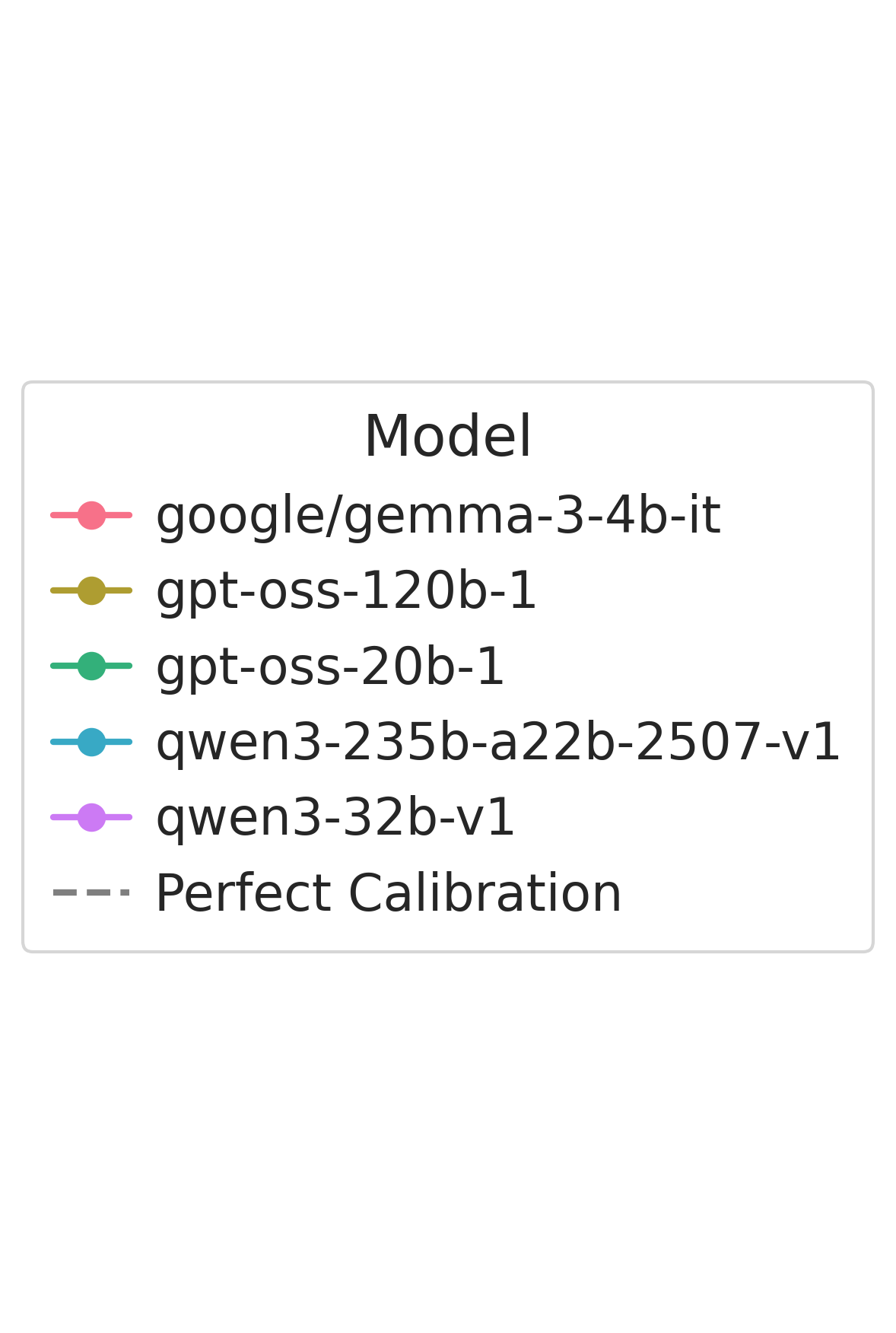}
    \end{minipage}
    \caption{\textbf{Self-reported confidence is poorly aligned with correctness.}
    Reliability diagrams across four verifier-absent benchmarks.
    Confidence increases faster than accuracy and does not reliably distinguish correct from incorrect answers.}
    \label{fig:confidence_calibration}
\end{figure*}

\subsection{Correlation Without Truth: A No-Signal Negative Control}
\label{sec:random_strings}

Correlation on benchmark questions could reflect shared knowledge rather than structural dependence between models. To disentangle these, we introduce a negative control. We generate 10{,}000 prompts, each consisting of a uniformly random sequence of 32 characters, with models forced to choose from $\{A,B,C,D\}$: 

\begin{tcolorbox}[colback=gray!5,colframe=gray!50,title=Random String Prompt]
\small\ttfamily
Here is a random sequence:\\ 

gP\%!mdq4k!'q=T/rp\textasciitilde j\textasciitilde LdW05[:Mkdk\$  \\

Now choose one option: (A), (B), (C), or (D). Output your answer as X where X is A, B, C, or D.
\end{tcolorbox}





Despite the absence of signal, model responses are not independent.
Across temperatures, several model pairs have a non-negligible positive correlation.
Moreover, the qualitative structure of inter-model correlation is stable across $T{=}0.0$ and $T{=}1.0$, indicating that temperature increases surface diversity without eliminating shared response biases.

Models exhibit correlated outputs even when no truth exists, confirming that correlation stems from shared inductive biases in model weights, not shared knowledge. Polling then amplifies these biases, increasing consensus without improving correctness.

\subsection{Consensus Is Easier Than Truth Verification}

Models predict collective opinion far better than they predict correctness.
Across benchmarks, vote-share predictions correlate strongly with vote shares, while confidence correlates weakly with accuracy.
This asymmetry reveals that confidence and predicted popularity track what the crowd will say, not whether the crowd is right.

\begin{figure}[t]
    \centering
    \includegraphics[width=\columnwidth]{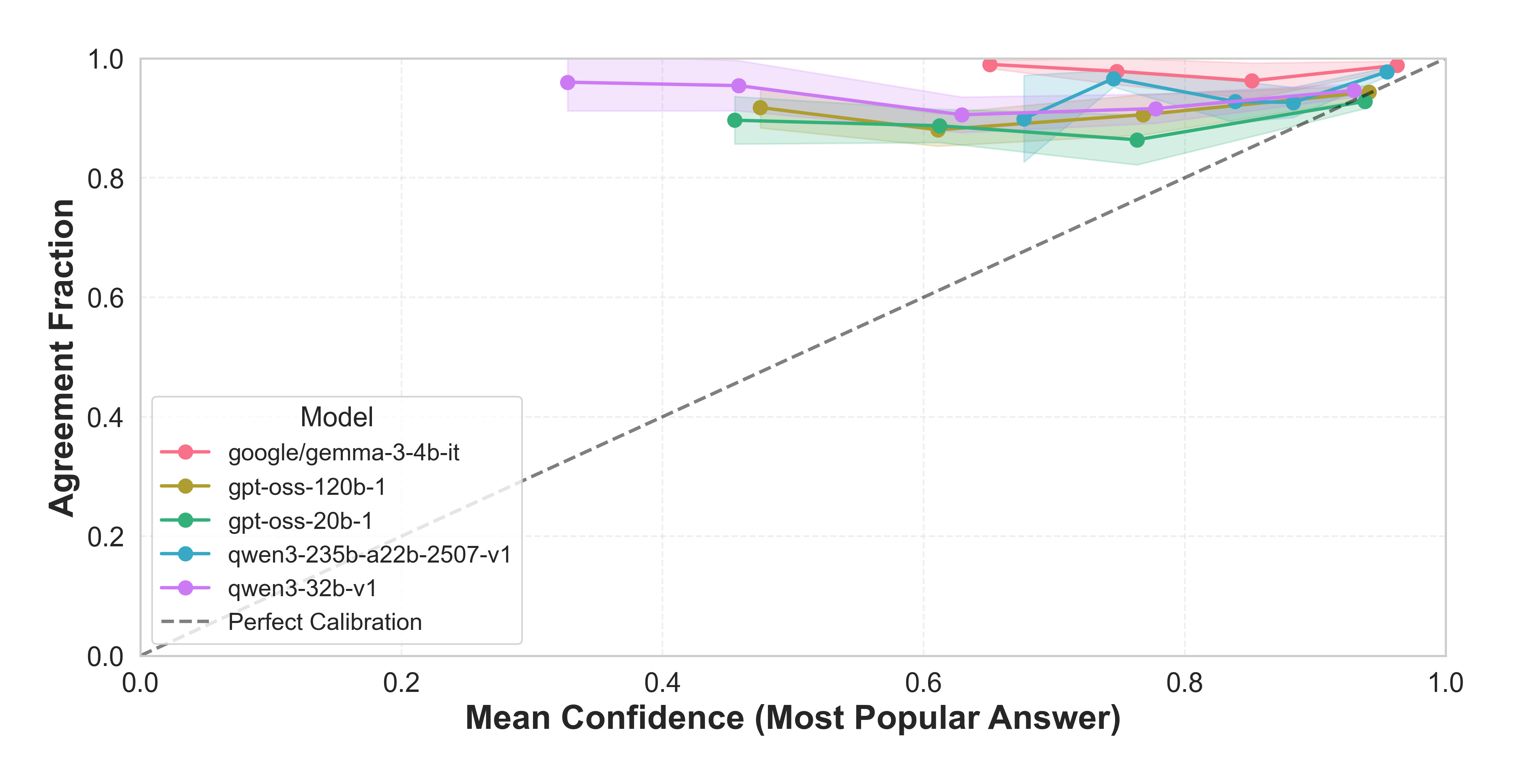}
    \caption{\textbf{Agreement remains high under low confidence.} Mean confidence for the most popular answer versus its agreement fraction. Deviations from $y = x$ show that consensus persists even as confidence decreases.}
    \label{fig:agreement_vs_confidence}
\end{figure}


\label{sec:diagnostics}

\begin{figure*}[t]
    \centering
    \begin{minipage}{0.22\textwidth}
        \centering
        \includegraphics[width=\linewidth]{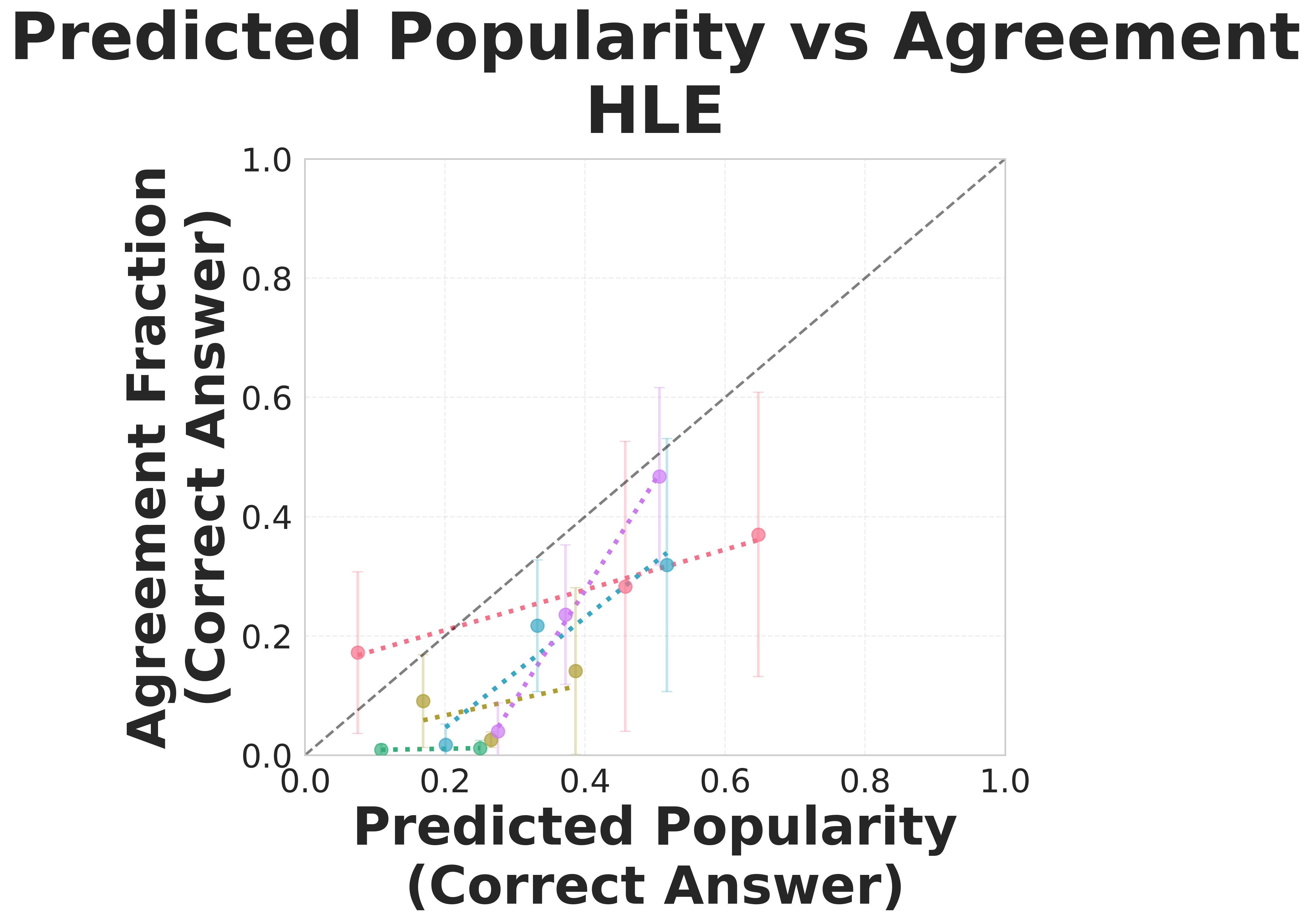}
        \subcaption{\textbf{HLE.}}
        \label{fig:predpop_hle}
    \end{minipage}
    \hfill
    \begin{minipage}{0.22\textwidth}
        \centering
        \includegraphics[width=\linewidth]{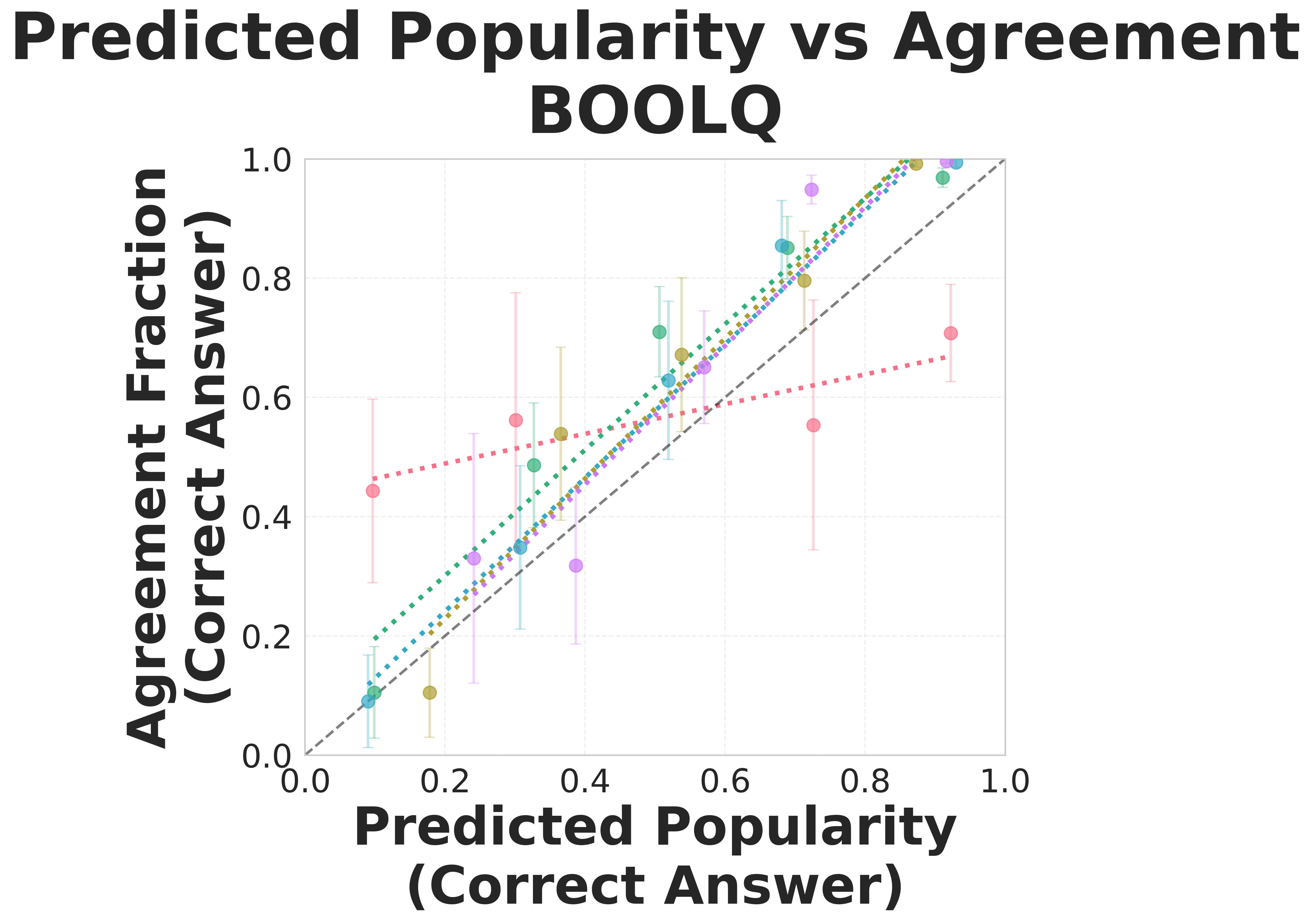}
        \subcaption{\textbf{BoolQ.}}
        \label{fig:predpop_boolq}
    \end{minipage}
    \hfill
    \begin{minipage}{0.22\textwidth}
        \centering
        \includegraphics[width=\linewidth]{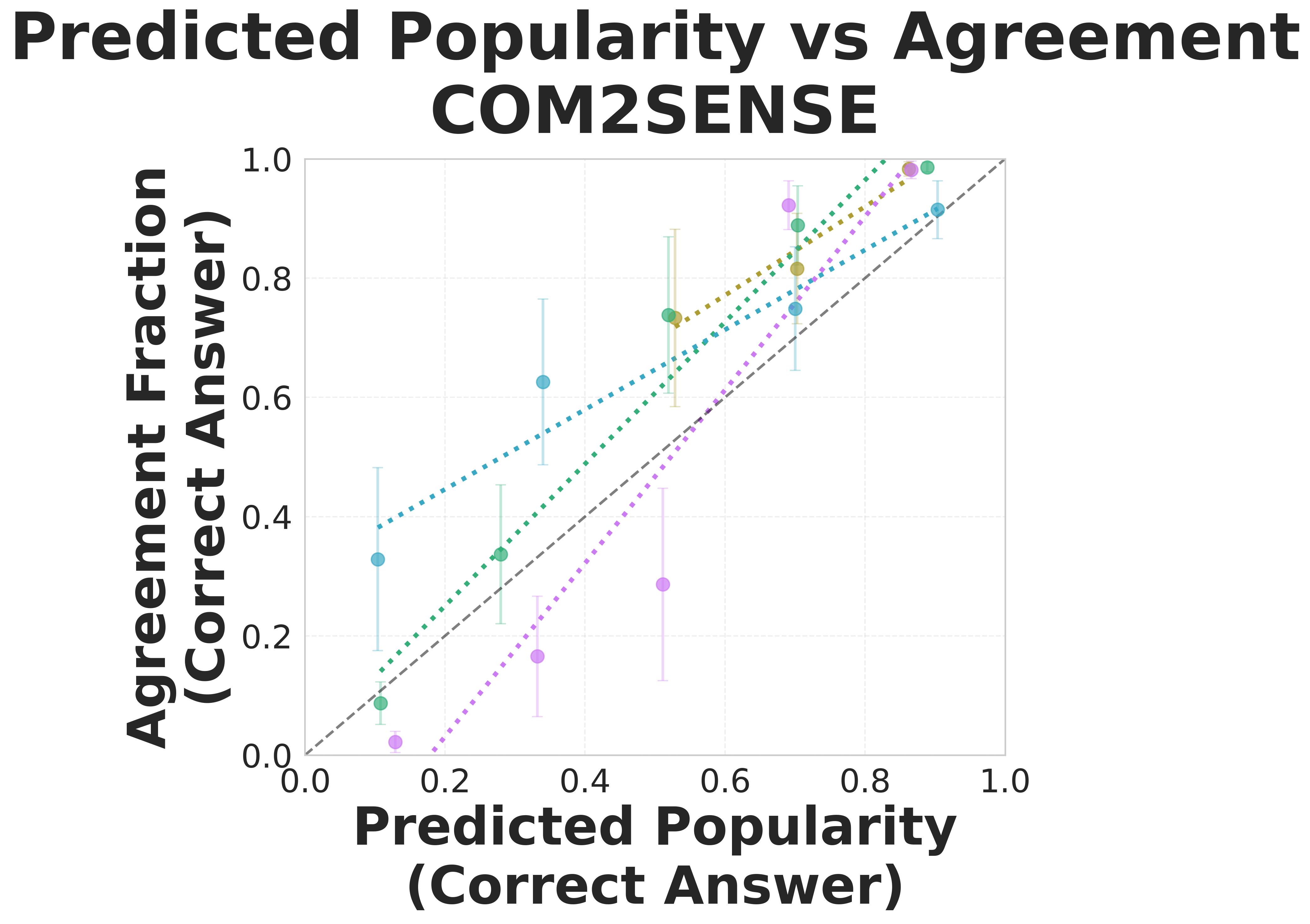}
        \subcaption{\textbf{Com2Sense.}}
        \label{fig:predpop_com2sense}
    \end{minipage}
    \hfill
    \begin{minipage}{0.22\textwidth}
        \centering
        \includegraphics[width=\linewidth]{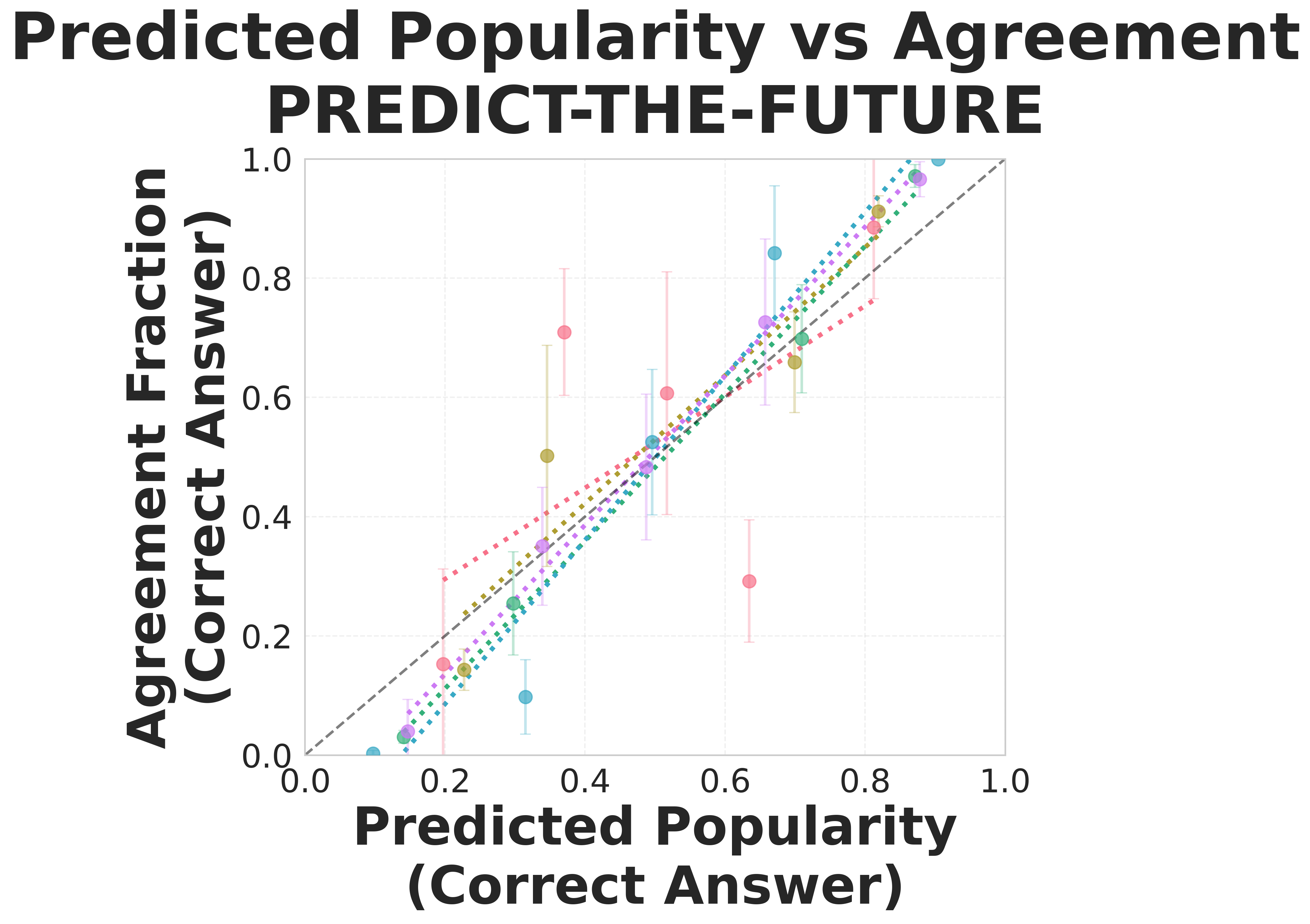}
        \subcaption{\textbf{Predict-the-Future.}}
        \label{fig:predpop_future}
    \end{minipage}
    \hfill
    \begin{minipage}{0.10\textwidth}
      \vspace{-40pt}  
      \includegraphics[width=\linewidth]{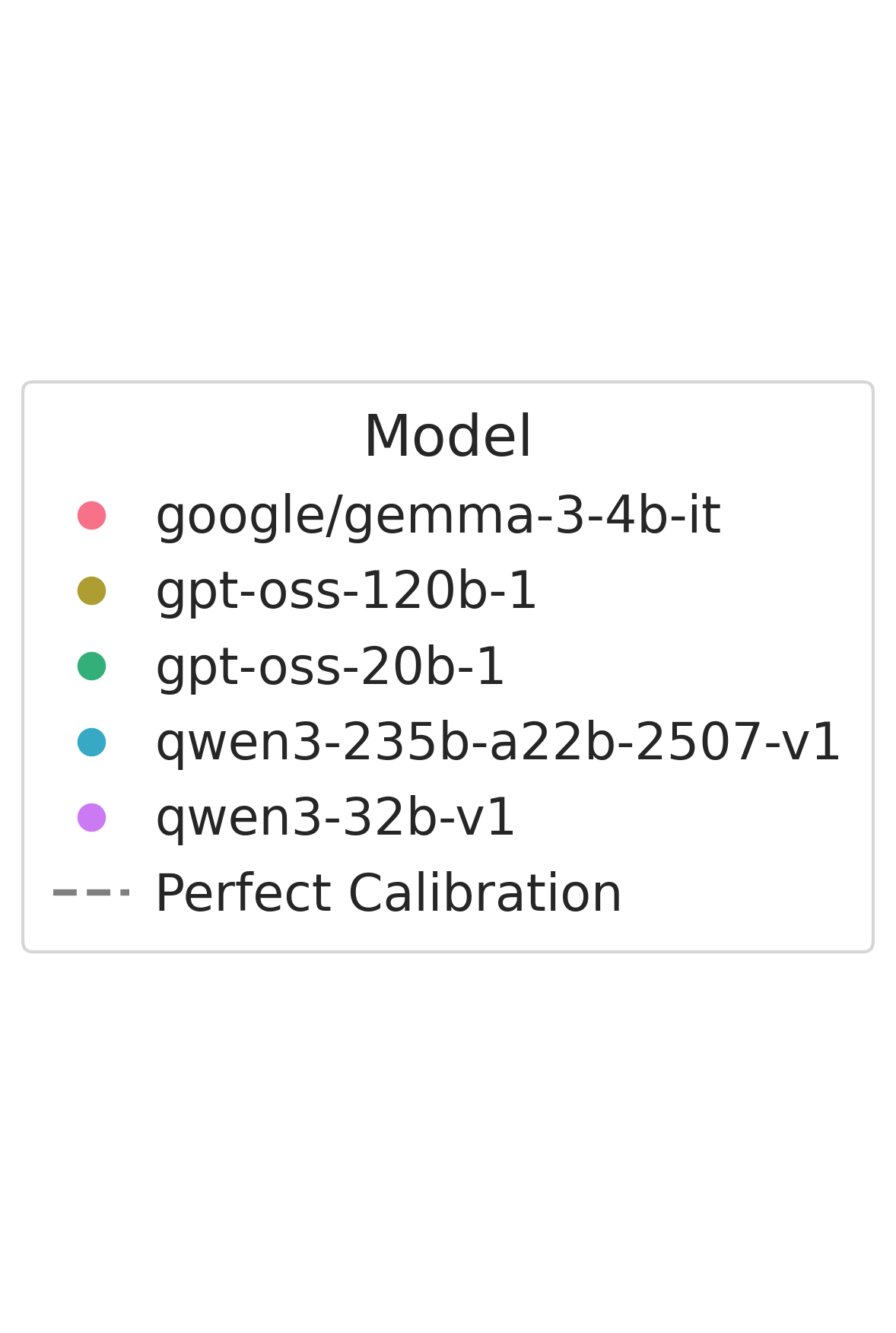}
    \end{minipage}

    \caption{\textbf{Predicted popularity tracks agreement, not truth.}
    Predicted vote share for the correct option (x-axis) versus observed agreement (y-axis).
    Points denote models; error bars show 95\% bootstrap confidence intervals.
    The dashed line indicates perfect calibration ($y=x$).
    Models accurately predict consensus even when consensus does not reliably indicate correctness.}
    \label{fig:predicted_popularity_vs_agreement}
\end{figure*}




\section{Discussion and Limitations}
\label{sec:discussion}

Our results identify a boundary condition for inference-time scaling.
Sampling and aggregation improve performance when candidates can be externally verified.
When no verifier exists, additional samples are typically drawn from the same epistemic prior.
In this regime, aggregation increases consensus faster than it increases truth.




\subsection{Why Internal Aggregation Signals Fail}

Richer internal signals do not resolve this limitation.
Agreement, confidence, predicted popularity, and surprise gaps are all strong predictors of what the crowd will say.
None provides a stable signal of correctness across tasks.

The Surprisingly Popular (SP) algorithm illustrates the problem.
Its success depends on a stable expert-minority structure.
Our diagnostics show that this structure is not reliably present in language model populations.
The sign of the surprise gap varies across datasets, and in some cases is anti-correlated with correctness.
A signal whose semantics change across tasks cannot function as a verifier.

Confidence-based methods fail for the same reason.
Verbalized confidence tracks expected agreement nearly as strongly as it tracks accuracy.
Weighting by confidence therefore amplifies the dominant misconception rather than correcting it.

\subsection{Social Prediction Is Easier Than Truth Verification}

Models predict collective opinion substantially better than they predict correctness.
This reveals a separation between two capabilities that are often conflated:
\emph{social prediction} (what will others say?) and \emph{truth verification} (is this answer correct?).
Most aggregation rules operate on the former.
When the crowd is systematically wrong, these signals become actively misleading.

This distinction suggests a constructive use of social prediction.
Rather than selecting answers, predicted consensus can be used to flag high-risk questions and trigger retrieval, tool use, or deferral.
In this framing, agreement is not evidence of truth, but a warning sign of shared failure modes.

\subsection{Limitations}

\paragraph{Binary response format.}
We restrict to binary questions to make vote shares and surprise-based aggregation well-defined.
This enables clean comparison across methods but limits direct applicability to open-ended generation.
Extensions to open-ended settings typically introduce implicit verification steps, which do not contradict our core claim.

\paragraph{Confidence elicitation.}
We use verbalized confidence to match black-box usage.
While alternative uncertainty estimates may behave differently, the central failure mode is not noise but misalignment: confidence tracks consensus nearly as strongly as correctness.

\paragraph{Diversity interventions.}
Temperature sampling and model ensembling change surface form more than beliefs.
Stronger interventions that induce genuine independence typically introduce new information, objectives, or supervision.
These effectively add verification or break shared priors, aligning with rather than refuting our conclusion.

\paragraph{Forecasting benchmark scope.}
Predict-the-Future is intentionally stringent but relatively small.
Larger-scale replications would strengthen external validity.

\subsection{What Would Change the Conclusion?}

Our results apply to \emph{self-aggregation without verification}.
Truthfulness may scale when systems:
\begin{enumerate}
    \item add grounding that functions as a verifier (retrieval, tools, execution, or human feedback),
    \item engineer genuine epistemic diversity through disjoint training or objectives, or
    \item learn explicit verifiers trained on externally labeled evidence.
\end{enumerate}

All three introduce information absent from pure polling.
Scaling truthfulness therefore requires verification or true independence, not more samples from a shared prior.

\section{Conclusion}

When language model errors are correlated, no aggregation rule based solely on agreement, confidence, or predicted popularity can reliably scale truthfulness without an external verifier.

Polling is not a substitute for verification. Across five instruction-tuned language models and multiple verifier-absent benchmarks, we find that agreement-based and metacognitive aggregation rules do not reliably improve accuracy over strong single-sample baselines, even at up to $25\times$ inference cost.

The failure is structural. Crowd wisdom relies on partially independent errors so that mistakes cancel under aggregation. Modern language models violate this assumption. Shared training data, objectives, and post-training incentives induce shared priors and shared blind spots, yielding strongly correlated errors. When the dominant answer is wrong, additional samples and additional models mostly reproduce the same mistake, turning aggregation into an amplifier of common misconceptions rather than a mechanism for correction.

This yields a clear corollary. \emph{When errors are correlated, no aggregation rule based solely on internal signals can reliably scale truthfulness in the absence of an external verifier.} Confidence, predicted popularity, and surprise gaps primarily track expected consensus, not correctness. Our diagnostics reveal a consistent separation between \emph{social prediction} and \emph{truth verification}: models are substantially better at forecasting what others will say than at identifying what is true. Signals that optimize the former cannot be repurposed into a verifier for the latter.

These results delineate a boundary for inference-time scaling. Aggregation improves performance when a verifier exists, because verification converts additional samples into additional correctness. In verifier-absent regimes, scaling truthfulness requires external grounding or interventions that actively break error correlation, rather than drawing more samples from a shared epistemic prior. In short, inference-time compute scales reasoning when verification is available, but it does not scale truth itself.

Our results have important implications for the future of language models.  Many problems, from reasoning to knowledge, have been solved simply by scaling up parameters, compute, and training data.  However, as models become more capable, the importance of eliciting truthfulness and obtaining reliable verifiers for scalable oversight will increase.  Our results imply that the naive approach of throwing compute at the problem is unlikely to suffice.  More creative solutions will be necessary to certify reliability and reinforce safety.

\clearpage 

\section*{Impact Statement}
We show that having multiple language models vote on answers does not reliably improve accuracy in domains that lack a verifier. Since models often make the same mistakes, agreement can be misleading. We hope this encourages more realistic expectations about what aggregation methods can achieve, and pushes toward approaches that actually verify correctness rather than simply measure consensus.

\clearpage

\bibliographystyle{icml2026}
\bibliography{references}

@misc{sutton2019bitter,
  author = {Sutton, Rich},
  title = {The Bitter Lesson},
  year = {2019},
  month = {March},
  day = {13},
  url = {http://www.incompleteideas.net/IncIdeas/BitterLesson.html}
}

@article{phan2025humanity,
  title   = {Humanity's Last Exam},
  author  = {Phan, Long and others},
  journal = {arXiv preprint arXiv:2501.14249},
  year    = {2025},
  url     = {https://arxiv.org/abs/2501.14249}
}

@misc{hendrycks2021measuringmathematicalproblemsolving,
      title={Measuring Mathematical Problem Solving With the MATH Dataset}, 
      author={Dan Hendrycks and Collin Burns and Saurav Kadavath and Akul Arora and Steven Basart and Eric Tang and Dawn Song and Jacob Steinhardt},
      year={2021},
      eprint={2103.03874},
      archivePrefix={arXiv},
      primaryClass={cs.LG},
      url={https://arxiv.org/abs/2103.03874}, 
}

@inproceedings{hughes2025bestofn,
title={Best-of-N Jailbreaking},
author={John Hughes and Sara Price and Aengus Lynch and Rylan Schaeffer and Fazl Barez and Arushi Somani and Sanmi Koyejo and Henry Sleight and Erik Jones and Ethan Perez and Mrinank Sharma},
booktitle={The Thirty-ninth Annual Conference on Neural Information Processing Systems},
year={2025},
url={https://openreview.net/forum?id=91l4ZTMpO4}
}

@inproceedings{schaeffer2025monkeys,
  title     = {How Do Large Language Monkeys Get Their Power (Laws)?},
  author    = {Schaeffer, Rylan and Kazdan, Joshua and others},
  booktitle = {International Conference on Machine Learning (ICML)},
  year      = {2025},
  note      = {Oral presentation},
  url       = {https://arxiv.org/abs/2502.17578}
}

@article{brown2024monkeys,
  title    = {Large Language Monkeys: Scaling Inference Compute with Repeated Sampling},
  author   = {Brown, Bradley and Juravsky, Jordan and Ehrlich, Ryan and Clark, Ronald and Le, Quoc V. and R{\'e}, Christopher and Mirhoseini, Azalia},
  journal  = {arXiv preprint arXiv:2407.21787},
  year     = {2024}
}

@inproceedings{huang2024large,
  title={Large Language Models Cannot Self-Correct Reasoning Yet},
  author={Huang, Jie and Chen, Xinyun and Mishra, Swaroop and Zhou, Denny and Yu, Zhou and Chi, Ed},
  booktitle={International Conference on Learning Representations (ICLR)},
  year={2024}
}

@article{kadavath2022language,
  title={Language Models (Mostly) Know What They Know},
  author={Kadavath, Saurav and Conerly, Tom and Askell, Amanda and Henighan, Tom and others},
  journal={arXiv preprint arXiv:2207.05221},
  year={2022}
}

@inproceedings{clark2019boolq,
  title={BoolQ: Exploring the Surprising Difficulty of Natural Yes/No Questions},
  author={Clark, Christopher and Lee, Kenton and Chang, Ming-Wei and Kwiatkowski, Tom and Collins, Michael and Toutanova, Kristina},
  booktitle={North American Chapter of the Association for Computational Linguistics (NAACL)},
  year={2019}
}

@inproceedings{singh2021com2sense,
  title={Com2Sense: A Commonsense Reasoning Benchmark with Complementary Sense Pairs},
  author={Singh, Shikhar and Ng, Nhan and Qiu, Lin and Thomas, Hanson and Ma, Tengyu},
  booktitle={Findings of the Association for Computational Linguistics (ACL)},
  year={2021}
}

@article{prelec2017solution,
  title={A Solution to the Single-Question Crowd Wisdom Problem},
  author={Prelec, Dra{\v{z}}en and Seung, H Sebastian and McCoy, John},
  journal={Nature},
  volume={541},
  number={7638},
  pages={532--535},
  year={2017}
}

@book{surowiecki2004wisdom,
  title={The Wisdom of Crowds},
  author={Surowiecki, James},
  year={2004},
  publisher={Anchor}
}

@article{vandolder2018wisdom,
  title={The wisdom of the inner crowd: A unified perspective},
  author={van Dolder, Dennie and van den Assem, Martijn J},
  journal={Management Science},
  volume={64},
  number={1},
  pages={457--473},
  year={2018}
}

@book{decondorcet1785essay,
  title={Essay on the Application of Analysis to the Probability of Majority Decisions},
  author={de Condorcet, Nicolas},
  year={1785},
  publisher={Imprimerie Royale}
}

@inproceedings{lakshminarayanan2017simple,
  title={Simple and Scalable Predictive Uncertainty Estimation using Deep Ensembles},
  author={Lakshminarayanan, Balaji and Pritzel, Alexander and Blundell, Charles},
  booktitle={Advances in Neural Information Processing Systems (NeurIPS)},
  year={2017}
}

@article{qwen2025qwen3,
  title={Qwen3 Technical Report},
  author={{Qwen Team}},
  journal={arXiv preprint arXiv:2505.09388},
  year={2025}
}

@article{openai2025gptoss,
  title   = {gpt-oss-120b \& gpt-oss-20b Model Card},
  author  = {Agarwal, Shreyas and others},
  journal = {arXiv preprint arXiv:2508.10925},
  year    = {2025},
  url     = {https://arxiv.org/abs/2508.10925}
}

@inproceedings{wang2023self,
  title={Self-Consistency Improves Chain of Thought Reasoning in Language Models},
  author={Wang, Xuezhi and Wei, Jason and Schuurmans, Dale and Le, Quoc and Chi, Ed and Narang, Sharan and Chowdhery, Aakanksha and Zhou, Denny},
  booktitle={International Conference on Learning Representations (ICLR)},
  year={2023}
}

@article{gemmateam2025gemma3,
  title   = {Gemma 3 Technical Report},
  author  = {{Gemma Team}},
  journal = {arXiv preprint arXiv:2503.19786},
  year    = {2025},
  url     = {https://arxiv.org/abs/2503.19786}
}

@article{qwen2024qwen,
  title={Qwen2.5 Technical Report},
  author={Yang, An and Yang, Baosong and Zhang, Binyuan and Hui, Biyuan and others},
  journal={arXiv preprint arXiv:2412.15115},
  year={2024}
}

@inproceedings{kim2025correlated,
  title={Correlated Errors in Large Language Models},
  author={Kim, Sung and others},
  booktitle={International Conference on Machine Learning (ICML)},
  year={2025}
}

@inproceedings{goel2025great,
  title={Great Models Think Alike and this Undermines AI Oversight},
  author={Goel, Shashank and others},
  booktitle={International Conference on Machine Learning (ICML)},
  year={2025}
}

@article{sharma2023sycophancy,
  title={Towards Understanding Sycophancy in Language Models},
  author={Sharma, Mrinank and Tong, Meg and Korbak, Tomasz and others},
  journal={arXiv preprint arXiv:2310.13548},
  year={2023}
}

@inproceedings{xiong2024can,
  title={Can LLMs Express Their Uncertainty? An Empirical Evaluation of Confidence Elicitation in LLMs},
  author={Xiong, Miao and Hu, Zhiyuan and Lu, Xinyang and Li, Yifei and Fu, Jie and He, Junxian and Hooi, Bryan},
  booktitle={International Conference on Learning Representations (ICLR)},
  year={2024}
}

@inproceedings{tian2023just,
  title={Just Ask for Calibration: Strategies for Eliciting Calibrated Confidence Scores from Language Models Fine-Tuned with Human Feedback},
  author={Tian, Katherine and Mitchell, Eric and Zhou, Allan and Sharma, Archit and Rafailov, Rafael and Yao, Huaxiu and Finn, Chelsea and Manning, Christopher},
  booktitle={Proceedings of the 2023 Conference on Empirical Methods in Natural Language Processing},
  pages={5433--5442},
  year={2023}
}

@inproceedings{leng2024taming,
  title={Taming Overconfidence in {LLMs}: Reward Calibration in {RLHF}},
  author={Leng, Jixuan and Huang, Chengsong and Zhu, Banghua and Huang, Jiaxin},
  booktitle={The Thirteenth International Conference on Learning Representations},
  year={2025},
  url={https://openreview.net/forum?id=l0tg0jzsdL}
}

@inproceedings{goodfellow2015explaining,
  title={Explaining and Harnessing Adversarial Examples},
  author={Goodfellow, Ian J and Shlens, Jonathon and Szegedy, Christian},
  booktitle={International Conference on Learning Representations (ICLR)},
  year={2015}
}

@inproceedings{snell2025scaling,
  title     = {Scaling {LLM} Test-Time Compute Optimally Can be More Effective than Scaling Parameters for Reasoning},
  author    = {Snell, Charlie Victor and Lee, Jaehoon and Xu, Kelvin and Kumar, Aviral},
  booktitle = {The Thirteenth International Conference on Learning Representations},
  year      = {2025},
  url       = {https://openreview.net/forum?id=4FWAwZtd2n}
}

@article{cohen1960coefficient,
  title={A coefficient of agreement for nominal scales},
  author={Cohen, Jacob},
  journal={Educational and Psychological Measurement},
  volume={20},
  number={1},
  pages={37--46},
  year={1960}
}

@misc{aime,
  title  = {American Invitational Mathematics Examination (AIME)},
  author = {{Mathematical Association of America}},
  year   = {1983--present}
}

\appendix

\section{The Surprisingly Popular Algorithm}
\label{app:sp}

The Surprisingly Popular (SP) algorithm \citep{prelec2017solution} addresses a fundamental limitation of democratic aggregation: majority voting fails when most respondents lack the relevant knowledge.
Standard approaches cannot distinguish between confident ignorance and genuine expertise.
SP solves this by eliciting predictions about others' responses, exploiting an information asymmetry between those who know and those who do not.

\paragraph{Setup.}
Consider a binary question with answers $A$ and $B$.
Each respondent $i$ provides:
\begin{enumerate}
    \item A vote $v_i \in \{A, B\}$
    \item A prediction $p_i \in [0,1]$ representing the expected fraction of respondents voting $A$
\end{enumerate}

Let $\bar{v}$ denote the actual fraction voting $A$, and $\bar{p}$ denote the average prediction.
The SP algorithm selects $A$ if $\bar{v} > \bar{p}$, and $B$ otherwise.
In words: choose the answer that is more popular than people predicted.

\paragraph{Intuition.}
The algorithm exploits an asymmetry in metacognitive awareness.
Consider the canonical example: ``Is Philadelphia the capital of Pennsylvania?''
Most people incorrectly believe yes (the majority is wrong).
Those who know the correct answer (Harrisburg) also know that Philadelphia is more famous and that most people will guess wrong.
When asked to predict the crowd, they forecast high support for Philadelphia.
Meanwhile, those who guess Philadelphia assume their answer is common and predict similarly.

The result: nearly everyone predicts high support for Philadelphia, but the actual vote share is lower because some respondents know the truth.
The ``no'' answer receives more votes than predicted, making it surprisingly popular.
This surprise reveals the presence of informed respondents whose knowledge would be drowned out by simple majority voting.

\paragraph{Worked example (Philadelphia).}
Consider the question ``Is Philadelphia the capital of Pennsylvania?'' with answers \texttt{YES} and \texttt{NO} (truth: \texttt{NO}, the capital is Harrisburg).
Suppose $80\%$ of respondents are uninformed and vote \texttt{YES}, while $20\%$ are informed and vote \texttt{NO}.
Each respondent also predicts the overall fraction voting \texttt{YES}$.$
Assume uninformed respondents predict $p_i=0.95$ (they expect near-consensus on \texttt{YES}), while informed respondents predict $p_i=0.80$ (they know most people will guess \texttt{YES}).

Let $\bar{v}_{\texttt{YES}}$ be the observed vote share for \texttt{YES} and $\bar{p}_{\texttt{YES}}$ the average predicted vote share for \texttt{YES}.
Then
\[
\bar{v}_{\texttt{YES}} = 0.80,
\qquad
\bar{p}_{\texttt{YES}} = 0.80\cdot 0.95 + 0.20\cdot 0.80 = 0.92.
\]
SP selects \texttt{YES} if $\bar{v}_{\texttt{YES}} > \bar{p}_{\texttt{YES}}$, and otherwise selects \texttt{NO}.
Here $\bar{v}_{\texttt{YES}} < \bar{p}_{\texttt{YES}}$, so SP selects \texttt{NO}.
Equivalently, the \texttt{NO} answer is ``surprisingly popular'' because its observed share exceeds its predicted share:
\[
\bar{v}_{\texttt{NO}} = 1-\bar{v}_{\texttt{YES}} = 0.20
\quad \text{and} \quad
\bar{p}_{\texttt{NO}} = 1-\bar{p}_{\texttt{YES}} = 0.08,
\]
so $\bar{v}_{\texttt{NO}} > \bar{p}_{\texttt{NO}}$.
Intuitively, informed respondents both (i) vote correctly and (ii) anticipate the majority's mistake, creating a systematic prediction gap that SP exploits.

\paragraph{Theoretical Guarantee.}
\citet{prelec2017solution} prove that under a Bayesian model where respondents share a common prior and update beliefs based on private signals, SP recovers the truth whenever the following conditions hold:
\begin{enumerate}
    \item Respondents report their beliefs honestly
    \item Respondents form unbiased predictions about others
    \item There exists a subpopulation with access to the correct answer
\end{enumerate}

The key insight is that truth-holders make systematically different predictions than non-truth-holders.
If $A$ is correct, those who know $A$ predict lower support for $A$ than those who believe $B$ (because $A$-knowers understand that $B$-believers exist and are common).
This prediction gap causes $A$ to exceed its predicted support, triggering selection by SP.

\paragraph{When SP Fails.}
The algorithm requires an ``expert minority'' structure: some respondents must know the truth while understanding that others do not.
SP fails when:
\begin{itemize}
    \item No respondent knows the answer (pure guessing)
    \item Everyone knows the answer (no surprise possible)
    \item Knowledgeable respondents cannot predict others' ignorance
    \item Errors are correlated such that wrong answers also exceed predictions
\end{itemize}

Our experiments test whether language models exhibit the expert-minority structure SP requires.
When they do (HLE), SP yields large gains.
When they do not (BoolQ), SP degrades accuracy by selecting answers that are surprisingly popular for reasons unrelated to correctness.

\section{Benchmark Construction}
\label{app:benchmark}

\paragraph{Datasets, splits, and filters.}
Table~\ref{tab:benchmark_construction} summarizes the exact splits and filters used in our experiments.

\begin{table*}[t]
\centering
\caption{Benchmark construction and filtering used in experiments.}
\label{tab:benchmark_construction}
\small
\begin{tabular}{lcccc}
\toprule
Dataset & Split & Filter/Mapping & Answer & \#Q \\
\midrule
HLE & test & exactMatch \& yes/no only & YES/NO & 35 \\
BoolQ & validation & bool $\rightarrow$ true/false & TRUE/FALSE & 100 \\
Com2Sense & validation & label $\rightarrow$ true/false & TRUE/FALSE & 100 \\
Predict-the-Future & train & normalize yes/no & YES/NO & 100 \\
\bottomrule
\end{tabular}
\end{table*}

\paragraph{Question counts and models.}
We use 35 HLE questions and 100 questions each for BoolQ, Com2Sense, and Predict-the-Future.

\paragraph{Sampling protocol summary.}
For each (question, model) we draw 25 samples at each temperature $T\in\{0.7,1.0\}$ for each experiment type (surprisingly popular and confidence-weighted). Each experiment issues both direct-answer prompts and prediction/confidence prompts, yielding 50 responses per (question, model, temperature, experiment). This totals 375{,}000 responses across all datasets.

\section{Single-Model Polling Results}
\label{app:single_model}

Tables~\ref{tab:per_model_hle_boolq} and~\ref{tab:per_model_future_com2sense} report per-model results at $T=1.0$.
Com2Sense omits Gemma-3-4B.

\begin{table*}[t]
\centering
\caption{Per-model results (T=1.0) on HLE and BoolQ. Values are accuracy with 95\% bootstrap CIs.}
\label{tab:per_model_hle_boolq}
\small
\setlength{\tabcolsep}{4pt}
\begin{tabular}{lcccccc}
\toprule
Model & Individual Avg. & Direct Majority & Highest Conf & Conf Weighted & Pred Weighted & Surp. Popular \\ \midrule
\multicolumn{7}{l}{\textbf{HLE} (T=1.0)} \\
Gemma-3-4B & 27.3\% [25.3, 29.4] & 28.7\% [14.3, 42.9] & 25.8\% [14.3, 40.1] & 28.6\% [14.3, 42.9] & 37.4\% [20.0, 51.5] & 28.7\% [14.3, 45.7] \\
GPT-OSS-20B & 10.1\% [8.3, 11.9] & 11.7\% [2.9, 22.9] & 12.0\% [3.0, 24.2] & 9.2\% [0.0, 21.2] & 8.5\% [0.0, 17.2] & 11.2\% [2.9, 22.9] \\
GPT-OSS-120B & 11.7\% [10.2, 13.3] & 8.7\% [0.0, 17.2] & 2.8\% [0.0, 8.6] & 8.4\% [0.0, 20.0] & 5.6\% [0.0, 14.3] & 8.4\% [0.0, 20.0] \\
Qwen3-32B & 28.2\% [26.2, 30.2] & 25.1\% [11.4, 40.0] & 22.9\% [8.6, 37.1] & 17.3\% [5.7, 28.6] & 19.6\% [5.7, 34.3] & 22.8\% [8.6, 37.1] \\
Qwen3-235B & 21.4\% [19.5, 23.4] & 20.2\% [8.6, 34.3] & 14.6\% [2.9, 28.6] & 14.6\% [2.9, 25.7] & 22.8\% [8.6, 37.1] & 25.4\% [11.4, 40.0] \\
\midrule
\multicolumn{7}{l}{\textbf{BoolQ} (T=1.0)} \\
Gemma-3-4B & 61.1\% [59.7, 62.4] & 61.0\% [51.0, 70.0] & 62.2\% [53.0, 72.0] & 62.1\% [52.0, 72.0] & 69.9\% [61.0, 79.0] & 60.1\% [50.0, 70.0] \\
GPT-OSS-20B & 74.0\% [72.7, 75.2] & 79.9\% [72.0, 88.0] & 74.4\% [65.0, 83.0] & 78.0\% [70.0, 86.0] & 66.9\% [57.0, 76.0] & 80.0\% [72.0, 87.0] \\
GPT-OSS-120B & 80.1\% [79.1, 81.3] & 80.0\% [72.0, 87.0] & 79.9\% [72.0, 88.0] & 84.0\% [76.0, 90.0] & 80.8\% [72.0, 88.0] & 82.0\% [75.0, 89.0] \\
Qwen3-32B & 74.8\% [73.7, 76.1] & 77.0\% [68.0, 85.0] & 74.0\% [65.0, 82.0] & 73.8\% [66.0, 82.0] & 75.2\% [66.0, 83.0] & 75.9\% [67.0, 84.0] \\
Qwen3-235B & 69.9\% [68.7, 71.2] & 69.9\% [61.0, 78.0] & 68.7\% [60.0, 77.0] & 71.2\% [63.0, 80.0] & 66.9\% [57.0, 76.0] & 72.1\% [62.0, 81.0] \\
\bottomrule
\end{tabular}
\end{table*}

\begin{table*}[t]
\centering
\caption{Per-model results (T=1.0) on Predict-the-Future and Com2Sense. Values are accuracy with 95\% bootstrap CIs.}
\label{tab:per_model_future_com2sense}
\small
\setlength{\tabcolsep}{4pt}
\begin{tabular}{lcccccc}
\toprule
Model & Individual Avg. & Direct Majority & Highest Conf & Conf Weighted & Pred Weighted & Surp. Popular \\ \midrule
\multicolumn{7}{l}{\textbf{Predict-the-Future} (T=1.0)} \\
Gemma-3-4B & 52.6\% [51.3, 53.8] & 53.8\% [44.0, 63.0] & 53.0\% [43.0, 63.0] & 52.0\% [43.0, 62.0] & 53.1\% [43.0, 63.0] & 52.1\% [43.0, 62.0] \\
GPT-OSS-20B & 50.1\% [48.5, 51.6] & 49.3\% [39.0, 59.0] & 51.4\% [42.4, 61.6] & 51.8\% [42.4, 61.6] & 50.7\% [41.0, 60.0] & 47.7\% [37.0, 58.0] \\
GPT-OSS-120B & 49.3\% [47.1, 51.3] & 45.4\% [34.7, 54.8] & 54.0\% [44.8, 63.5] & 46.9\% [36.5, 56.2] & 51.8\% [42.0, 61.0] & 50.4\% [41.1, 61.1] \\
Qwen3-32B & 49.8\% [48.4, 51.1] & 50.0\% [40.0, 60.0] & 48.8\% [39.0, 58.0] & 48.1\% [38.0, 58.0] & 41.0\% [32.0, 50.0] & 50.1\% [40.0, 60.0] \\
Qwen3-235B & 51.8\% [50.4, 53.1] & 52.1\% [42.0, 61.0] & 55.2\% [45.0, 65.0] & 53.8\% [44.0, 63.0] & 48.9\% [39.0, 59.0] & 52.0\% [42.0, 62.0] \\
\midrule
\multicolumn{7}{l}{\textbf{Com2Sense} (T=1.0)} \\
GPT-OSS-20B & 62.5\% [61.5, 63.4] & 38.9\% [29.0, 49.0] & 40.2\% [31.0, 50.0] & 36.9\% [28.0, 46.0] & 38.0\% [29.0, 48.0] & 45.0\% [35.0, 55.0] \\
GPT-OSS-120B & 65.0\% [64.0, 65.9] & 41.2\% [31.0, 51.0] & 38.2\% [28.0, 47.0] & 41.1\% [31.0, 51.0] & 39.9\% [30.0, 50.0] & 45.8\% [36.0, 56.0] \\
Qwen3-32B & 51.7\% [50.7, 52.7] & 28.3\% [20.0, 37.0] & 40.9\% [31.0, 50.0] & 34.0\% [24.0, 44.0] & 30.0\% [21.0, 39.0] & 32.8\% [23.0, 42.0] \\
Qwen3-235B & 75.6\% [74.3, 76.8] & 77.1\% [69.0, 85.0] & 72.9\% [64.0, 81.0] & 72.2\% [63.0, 80.0] & 69.0\% [60.0, 78.0] & 75.3\% [67.0, 84.0] \\
\bottomrule
\end{tabular}
\end{table*}

\begin{figure*}[t]
\centering

\begin{subfigure}{0.95\textwidth}
\centering
\includegraphics[width=\linewidth]{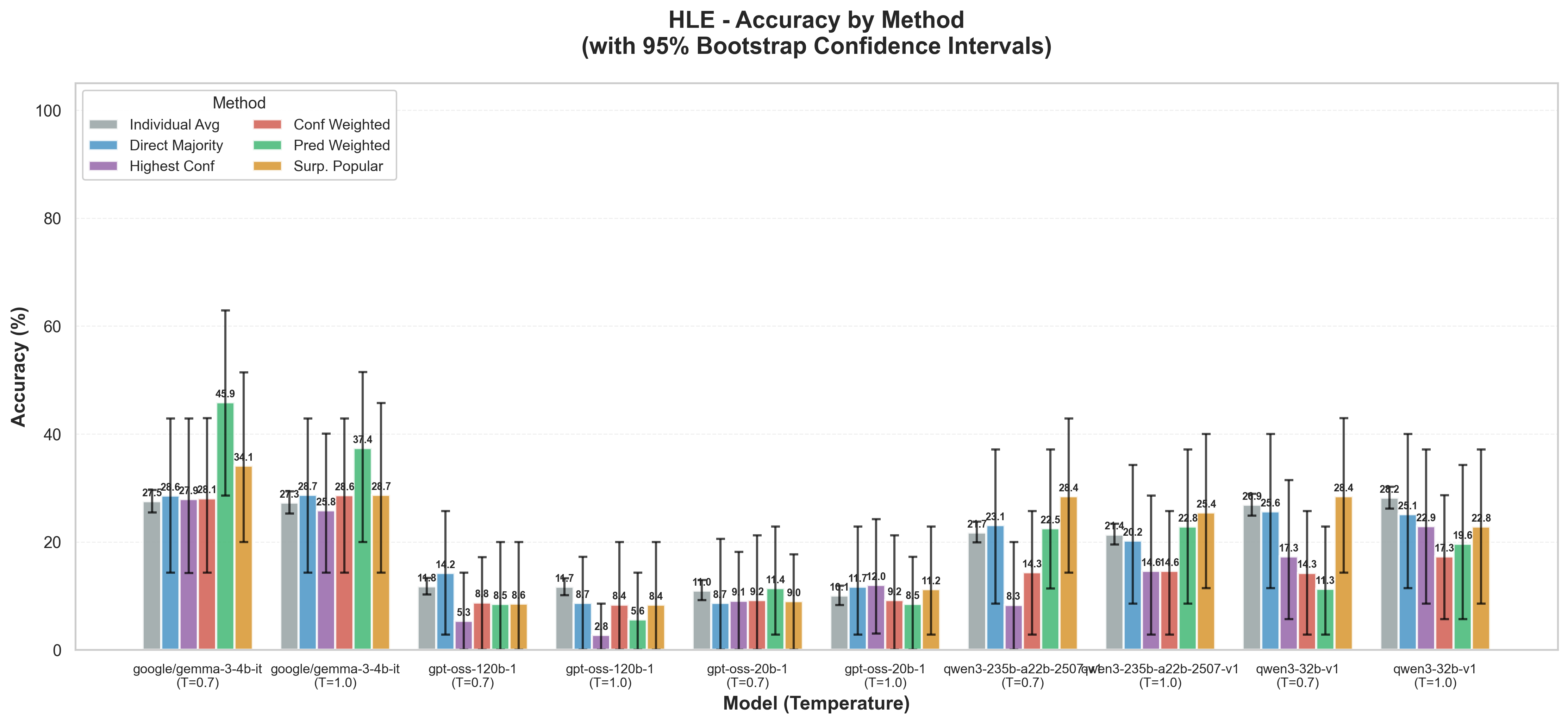}
\subcaption{HLE (binary subset).}
\label{fig:hle_per_model}
\end{subfigure}

\vspace{0.75em}

\begin{subfigure}{0.95\textwidth}
\centering
\includegraphics[width=\linewidth]{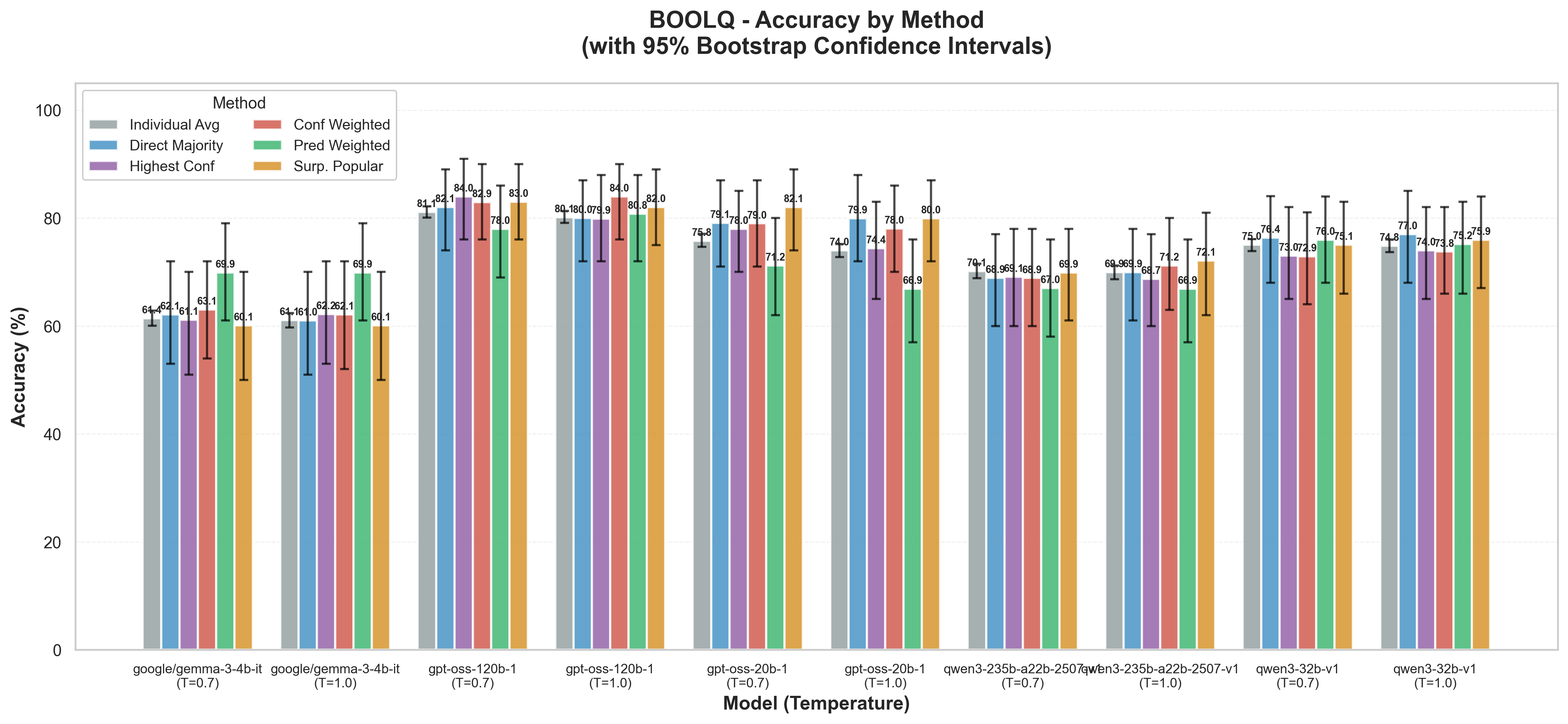}
\subcaption{BoolQ.}
\label{fig:boolq_per_model}
\end{subfigure}

\caption{\textbf{Per-model aggregation accuracy on factual and expert benchmarks.}
Accuracy by aggregation rule at temperatures $T\in\{0.7,1.0\}$. Bars correspond to individual-sample average, direct majority vote, highest-confidence selection, confidence-weighted vote, predicted-popularity weighting, and Surprisingly Popular. Error bars show 95\% bootstrap confidence intervals over questions.}
\label{fig:per_model_accuracy_main}
\end{figure*}

\begin{figure*}[t]
\centering

\begin{subfigure}{0.95\textwidth}
\centering
\includegraphics[width=\linewidth]{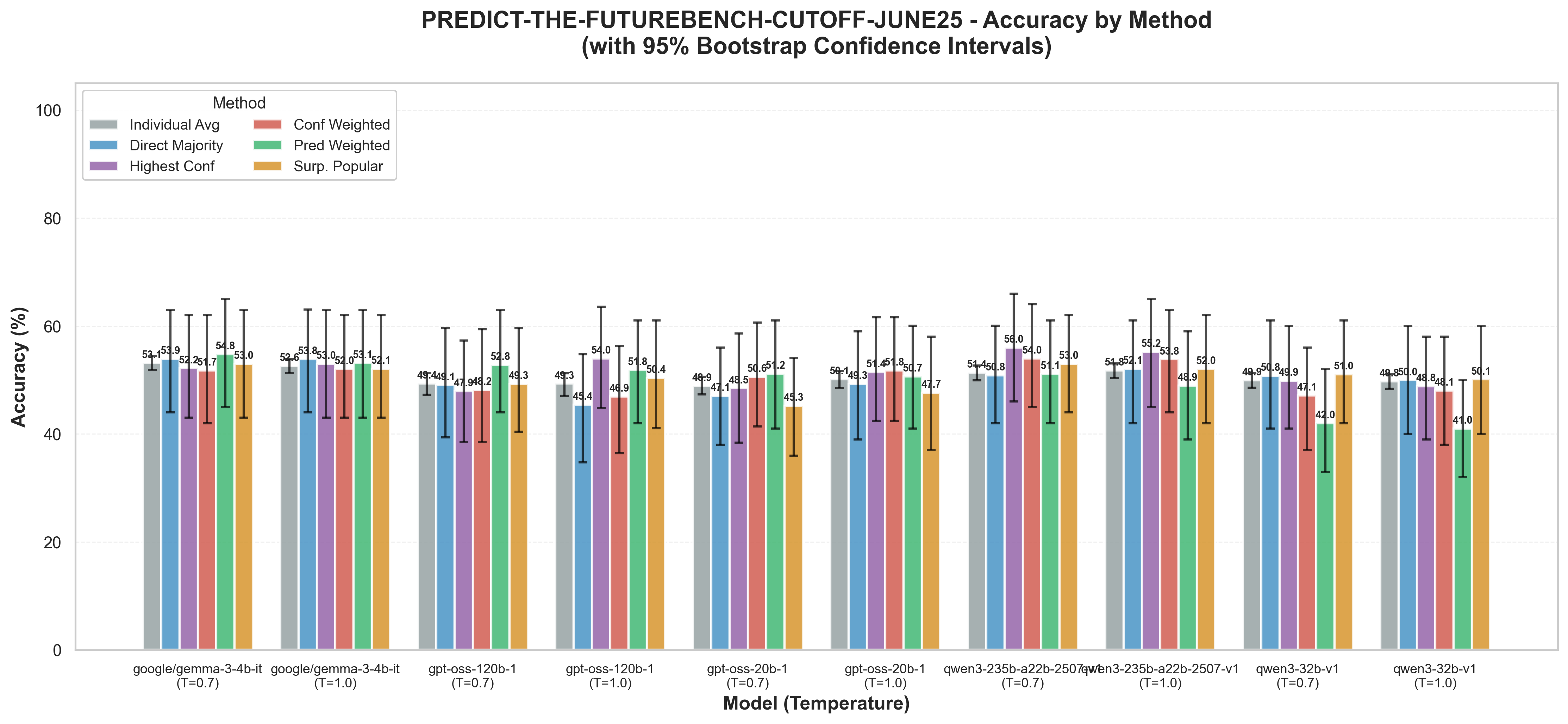}
\subcaption{Predict-the-Future (post-cutoff forecasting).}
\label{fig:future_per_model}
\end{subfigure}

\vspace{0.75em}

\begin{subfigure}{0.95\textwidth}
\centering
\includegraphics[width=\linewidth]{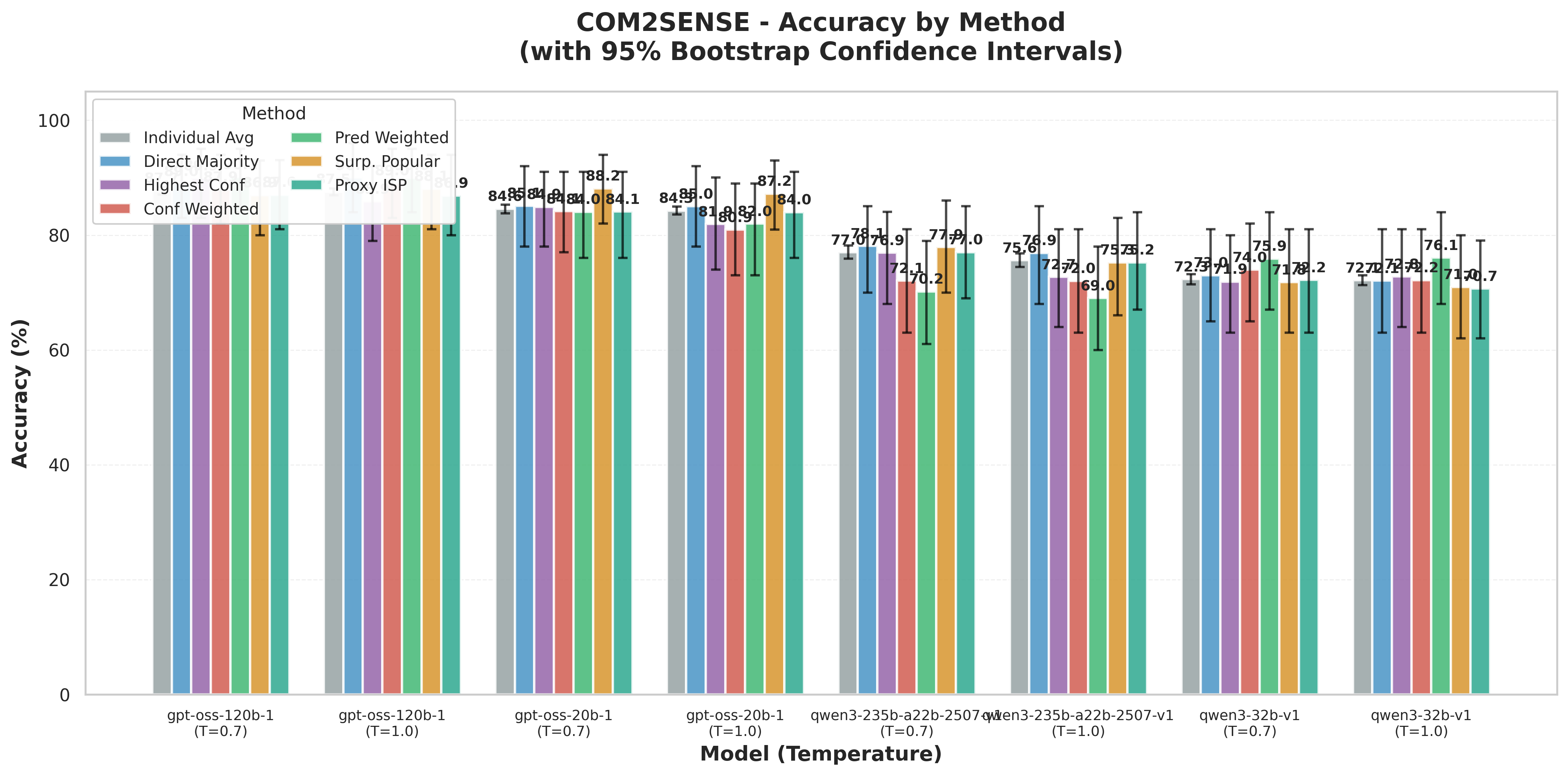}
\subcaption{Com2Sense.}
\label{fig:com2sense_per_model}
\end{subfigure}

\caption{\textbf{Aggregation fails to extract truth in verifier-absent regimes.}
Forecasting and commonsense benchmarks show no consistent gains from aggregation over single-sample baselines, even as agreement increases. Error bars show 95\% bootstrap confidence intervals.}
\label{fig:per_model_accuracy_aux}
\end{figure*}

\section{Prompts and Parsing Details}
\label{app:prompts}

\paragraph{Base prompts.}
We use the following templates for binary questions. For TRUE/FALSE tasks, replace YES/NO with TRUE/FALSE.

\begin{tcolorbox}[colback=gray!5,colframe=gray!50,title=Direct Answer]
\small\ttfamily
<Question>\newline
Answer with just YES or NO:
\end{tcolorbox}

\begin{tcolorbox}[colback=gray!5,colframe=gray!50,title=Surprisingly Popular (Prediction)]
\small\ttfamily
Question: <Question>\newline
What percentage of language models will answer YES to this question? Give just a number 0--100.\newline
What percentage will answer NO? (should sum to 100)\newline
Format: YES:XX NO:XX
\end{tcolorbox}

\begin{tcolorbox}[colback=gray!5,colframe=gray!50,title=Confidence]
\small\ttfamily
Question: <Question>\newline
Answer YES or NO, then rate your confidence from 0--100.\newline
Format: Answer:YES Confidence:XX or Answer:NO Confidence:XX
\end{tcolorbox}

\paragraph{Dataset-specific variants.}

\begin{tcolorbox}[colback=gray!5,colframe=gray!50,title=Com2Sense (Direct)]
\small\ttfamily
<Scenario>\newline
Does this scenario make sense? Answer with just TRUE (makes sense) or FALSE (doesn't make sense):
\end{tcolorbox}

\begin{tcolorbox}[colback=gray!5,colframe=gray!50,title=Predict-the-Future (Direct)]
\small\ttfamily
<Question>\newline
You may not know the answer for certain, but please make your best prediction. Answer with just YES or NO:
\end{tcolorbox}

For Com2Sense, the SP and confidence prompts use the same wording with TRUE/FALSE substitutions.  
For Predict-the-Future, the SP prompt adds the line “Note: The answer may not be knowable for certain, but predict based on available information.”

\paragraph{Parsing rules.}
Direct answers are extracted by lowercasing the response and checking for the presence of the target options (YES/NO or TRUE/FALSE). If neither option is found, the response is labeled \texttt{unclear} and excluded from vote counts. Majority ties default to the second option (NO or FALSE).

For Surprisingly Popular predictions, we parse numbers using regex patterns \texttt{YES:?\textbackslash s*(\textbackslash d+)} (or \texttt{TRUE:?\textbackslash s*(\textbackslash d+)}). Missing values default to 0.5. If both option counts are zero, the observed vote rate defaults to 0.5.

For confidence responses, we parse \texttt{Answer:?\textbackslash s*(yes|no|true|false)} and \texttt{Confidence:?\textbackslash s*(\textbackslash d+)}. Missing confidence defaults to 0.5; missing answers are treated as \texttt{unclear} and ignored in confidence-weighted voting.

\section{Temperature Stability Analysis}
\label{app:temperature}

To test whether temperature variation induces meaningful answer diversity, we compared \emph{plurality} answers between $T{=}0.7$ and $T{=}1.0$ for each (benchmark, model, question) tuple.
Table~\ref{tab:temp_stability} reports the fraction of cases in which the plurality answer flips.

\begin{table}[ht]
  \centering
  \caption{Plurality answer flip rate between $T{=}0.7$ and $T{=}1.0$. Low flip rates indicate that temperature variation does not induce the error independence required for effective aggregation.}
  \label{tab:temp_stability}
  \small
  \begin{tabular}{@{}lcc@{}}
    \toprule
    Benchmark & Flip Rate & Questions $\times$ Models \\
    \midrule
    Predict-the-Future & 3.8\% & 500 \\
    BoolQ              & 3.2\% & 500 \\
    HLE                & 2.3\% & 175 \\
    Com2Sense          & 1.5\% & 400 \\
    \midrule
    \textbf{Overall}   & \textbf{2.9\%} & 1,575 \\
    \bottomrule
  \end{tabular}
\end{table}

Flip rates vary by model: GPT-OSS-20B shows the highest instability (6.6\%), while Gemma-3-4B is nearly deterministic (0.4\%). GPT-OSS-20B on both BoolQ and Predict-the-Future reaches 9.0\%, the highest in our evaluation. These patterns suggest smaller models may be more sensitive to temperature, though all models remain far below the diversity levels that would enable effective crowd aggregation.

\end{document}